\documentclass[a4paper,11pt,onecolumn]{scrartcl}

\usepackage{graphicx}
\usepackage{amsmath}
\usepackage{amssymb}
\usepackage{nicefrac}
\usepackage{pifont}

\usepackage[pdftex,
            pdfauthor={Dominik Alexander Klein, Boris Illing, Bastian Gaspers, Dirk Schulz, Armin Bernd Cremers},
            pdftitle={Hierarchical Salient Object Detection for Assisted Grasping},
            pdfsubject={Computer Vision for Robotics},
            pdfkeywords={},
            pdfproducer={},
            pdfcreator={},
            pagebackref=true,breaklinks=true,letterpaper=false,colorlinks,bookmarks=false]{hyperref}
\usepackage{fancyhdr}

\newcommand{\cmark}{\ding{51}}%
\newcommand{\xmark}{\ding{55}}%

\begin{document}
\pagestyle{fancy}
\lhead{Accepted for ICRA 2017}
\rhead{Final version will be available at \url{ieeexplore.ieee.org}}

\title{Hierarchical Salient Object Detection for Assisted Grasping} 

\author{Dominik Alexander Klein$^1$, Boris Illing$^1$, Bastian Gaspers$^1$,\\ Dirk Schulz$^1$, and Armin Bernd Cremers$^2$\\
\begin{small}
 $^1$ Fraunhofer FKIE, Dept. Cognitive Mobile Systems
\end{small}\\
\begin{small}
 $^2$ Bonn-Aachen Int. Center for Information Technology (b-it)
\end{small}}
\date{January 17, 2017}

\maketitle

\begin{abstract}
 Visual scene decomposition into semantic entities is one of the major challenges
 when creating a reliable object grasping system. Recently, we
 introduced a bottom-up hierarchical clustering approach~\cite{Klein2016} which
 is able to segment objects and parts in a scene. In this paper, we introduce a
 transform from such a segmentation into a corresponding, hierarchical saliency
 function. In comprehensive experiments we demonstrate its ability to detect
 salient objects in a scene. Furthermore, this hierarchical saliency defines
 a most salient corresponding region (scale) for every point in an image. Based on this,
 an easy-to-use pick and place manipulation system was developed and tested exemplarily.
\end{abstract}

\section{Introduction and Related Work}
\begin{figure}
   \begin{center}
    \includegraphics[width=0.7\linewidth]{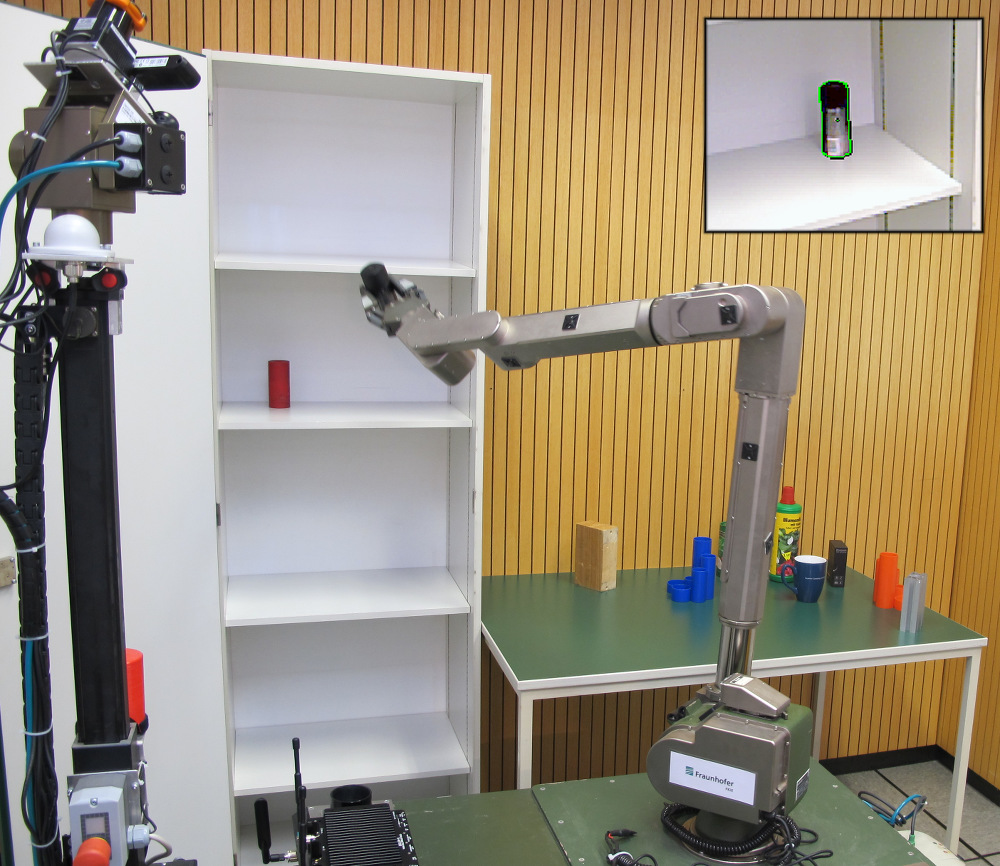}
   \end{center}
   \caption{Mobile robot picking a cylindrical object out of a cabinet. In the top-right corner an example of the salient object segmentation is shown, highlighting the object before execution of the grasp.}
   \label{fig:teaser}
  \end{figure}

 Intelligent robots need to comprehend the configuration of their environment in
 order to interact with it. We propose to follow a biologically motivated pathway,
 mimicking visual grouping mechanisms to decompose the scene and saliency to assess
 what the important entities are. Instead of implementing these two abilities
 independently, we show how to smartly transform the cognizance of grouping
 to awareness of salient structures. In the human mind, these mechanisms both belong to
 the unconscious, early sensory input processing, prior to recognition and high level
 semantic interpretation. On that basis, we built a framework for assisted object
 grasping with a mobile platform, that means an operator adds the high-level knowledge. Findings from research on the benefit of human input in human-in-the-loop robotic systems for pick and place tasks~\cite{bringes2013determining} indicate that overall performance is maximized when an operator specifies an approximate region of grasping without full control of the manipulator movement.
 Our hierarchical salient object detection approach facilitates a convenient way for an operator to select and segment an object of interest. Based on this,
 we created a flexible and easy-to-use grasping framework (see Fig.~\ref{fig:teaser}).
 Ideally, a pick and place task for an object is performed with only a single
 click each for selection, point of application, and target position.

 For hierarchical grouping we merely employ our previous work from~\cite{Klein2016}
 and thus focus on saliency based object detection in this paper. There is a large
 body of literature on visual saliency systems. Recently, Borji et.~al.~\cite{SalObjBenchmark}
 conducted a comprehensive, comparative study. The most successful approaches all
 conduct a kind of clustering prior to saliency estimation. Several ones apply a
 Superpixel algorithm as preprocessing on a single~\cite{perazzi2012,RBD,CHM}
 or multiple scales~\cite{DRFI,ZhuKlein2014}, whereupon the Superpixels are embedded in
 a graph structure to direct contrast based saliency computations.
 The approach of Aytekin et.~al.~\cite{QCUT} is special in the way that they derive
 a saliency value from a Graph-Cut like energy minimization to separate fore- and
 background.
 Most similar to our saliency approach, there are the ones based on hierarchical
 clusterings~\cite{HS,ST,HARF}. Still, the approach of Yan et.~al.~\cite{HS} differs
 a lot in the way that it extracts predefined planar scale levels from the hierarchical
 segmentation instead of directly exploiting the partition tree structure.
 Zhu and Komodakis~\cite{HARF} make use of a hierarchical clustering for subsequent,
 informed feature extraction. They trained a classifier on the ground truth labels of
 a fraction of the MSRA10K benchmark dataset, which is finally applied to grade
 saliency of segments within the partition tree structure.
 While Liu et.~al.~\cite{ST} actually utilize the hierarchical clustering for a likewise
 hierarchical saliency calculation as we do, our approach differs, since we use
 the same similarity measure for grouping as well as saliency. Additionally, we harness
 the present information about the neighborhood scale at the time of merging for
 foreground/background distinction and suggest how saliency is inherited from parent
 to children nodes in the partition tree. Moreover, we propose a maximum projection in the
 tree structure to define a saliency map. This retains the possibility to attribute
 the most salient segment's extent as the appropriate scale of a pixel.

The field of robotic manipulation can typically be divided into solutions for grasping of known objects or more general approaches for manipulating unknown objects. A plethora of well working solutions for the first category have already been proposed, for example the MOPED framework for the robotic butler HERB~\cite{collet2011moped}. When considering unknown objects, which is still an unsolved problem for general and reliable grasping, there exist several strategies for the segmentation.
Frameworks with saliency based object detection in a pick and place context typically focus on full autonomy by choosing the most salient object for grasping. For instance, the approach introduced by Bao et.~al.~\cite{bao2015saliency}
provides such level of autonomy to specify high-level instructions like "clean up the floor" to a mobile robot. The downside of these solutions, however, is that the operator usually cannot adjust the salient segmentation  and resulting manipulation actions.
Contrary to the automated saliency based approaches, there exist many variants of assisted pick and place with RGB-D cameras, which solely rely on depth information for segmenting unknown objects. Usually, these approaches typically detect a support plane and use a threshold pattern to cluster depth points above the plane that belong to an object selected by an operator. A prominent example for this is the work by Willow Garage on the PR2~\cite{ciocarlie2014towards}. A comparable variant was also successfully used for competition runs at RoboCup@Home~\cite{stuckler2013efficient}. These solutions work well for objects that are spatially well distinguishable from each other, but fail for crowded scenes, stacked objects, or a missing support plane. These downsides, however, can successfully be circumvented through our hierarchical saliency based object segmentation.

\section{Hierarchical Saliency from Segmentation}
 \begin{figure*}[t]
  \begin{center}
   \includegraphics[width=\linewidth]{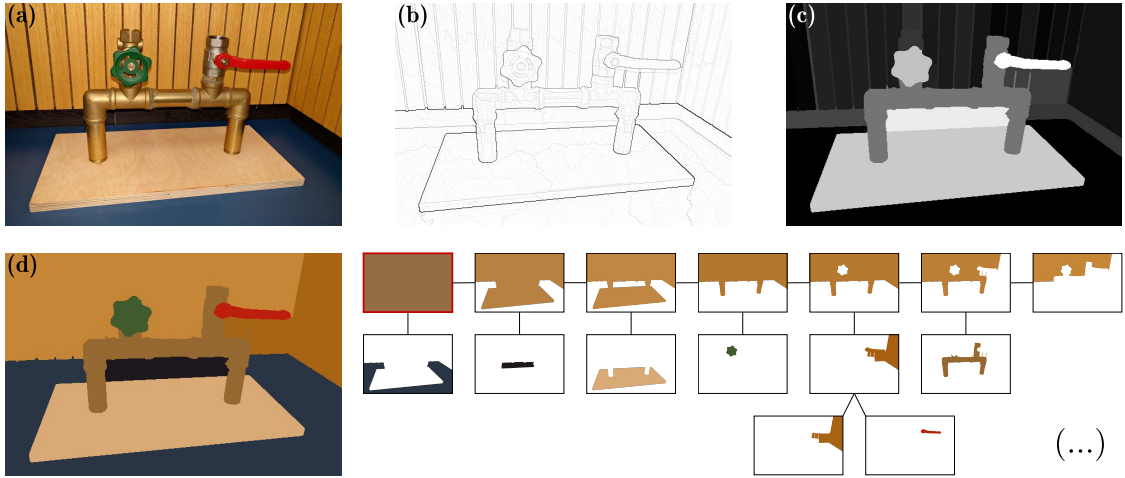}
  \end{center}
  \caption{(a) Input image. (b) Ultrametric contour map (UCM) visualizing $\mathcal{D}(P,Q)$ from~\cite{Klein2016}. (c) Proposed PaTS model. (d) Partitioning and corresponding tree for the first seven binary splits. Note how relatively smaller segments enclosed by strong contours become salient.}
  \label{fig:part_tree}
 \end{figure*}
   
 In our approach, the saliency computation is deeply geared with our hierarchical
 segmentation from~\cite{Klein2016}. It is calculated solely on the information
 already contained in the binary partition tree structure resulting from that
 clustering. However, the principle how to transform a hierarchical clustering to
 a saliency function is generic and could be applied to different approaches as
 well.
 In the following, we first give a brief summary on our hierarchical clustering
 procedure and which data is important to keep track of. Then, we show how to
 transform this information into a saliency value for each segment in the partition
 tree. Finally, we explain how to render a single saliency map from this hierarchical
 saliency function. 

\subsection{Growing the Scene Partitioning Tree}
 Our approach follows the traditional scheme of agglomerative clustering, starting
 from an atomic partitioning generated from a color-gradient based watershed
 transform. Then, iteratively, those two segments $P$ and $Q$ neighboring in image
 space become merged into a parent node, which are globally most similar with
 respect to a sophisticated distance measure $\mathcal{D}(P,Q)$ (cf.~\cite{Klein2016}, Eq.~(2)).
 For saliency computations, we only annotate the areas $A_P$ resp. $A_Q$ (in pixels),
 the length of the perimeter touching the image boundary (where applicable),
 as well as their distance at the time of merging to the binary partition tree.
 The agglomerative clustering stops when the last two segments are merged to
 become root, representing the full scene.

\subsection{Saliency Transform from Composition and Contrast}
 We claim that saliency and grouping are the same fundamental process, but considered
 from two opposing viewpoints: in grouping, one focuses on which parts of the image
 are most similar and should belong together, whereas saliency tells us which ones
 are most contrasting and should stay separate. This motivated the idea to utilize the
 information already gained during hierarchical clustering to express an antagonal
 saliency.
 
 There are some important questions to answer when composing a contrast based
 saliency term from the hierarchical grouping similarities: first of all, which of the
 two opposing segments is foreground and which one is background? Second, how should the
 image boundary be treated, could one model effects of human peripheral vision to
 saliency?  Finally, how is saliency inherited from object scale down to parts?
 Surprisingly, we found straightforward and well-performing solutions to all of
 these issues. Since our approach is working purely on the graph structure created
 from hierarchical grouping, we call it \textit{Partition Tree Saliency} (PaTS).
 Figure~\ref{fig:part_tree} shows an example of the hierarchical grouping and the
 corresponding saliency.
 
 The figure-ground relationship is decided inverse proportionally to the size of
 the segments at time of merging: the larger one is more likely to be background
 and therefore less salient. This yields
 \begin{equation}
  \mathcal{S}_{\text{fg}}(P)_Q = \mathcal{D}(P,Q) \cdot \frac{A_Q}{A_P + A_Q} .
 \end{equation}
 Subsequently, this measure is weighted down to model the effect of peripheral vision,
 assuming an expansion beyond the image boundary as
 \begin{equation}
  \mathcal{S}_{\text{peri}}(P)_Q = \mathcal{S}_{\text{fg}}(P)_Q \cdot \frac{A_P}{A_P + \nicefrac{(B_P^2 + \sqrt{A_P} B_P)}{2}} ,
 \end{equation}
 where $B_P$ denotes the length of that perimeter of segment $P$ running along the
 image boundary. This factor vanishes to $1$, if there is no image boundary involved.
 Otherwise it decreases with the average of squared plus linear dependence on the
 boundary length relative to the segment's surface. A different variant of such mechanism, called
 \textit{boundary connectivity}, was introduced in~\cite{RBD}.\\
 Lastly, we propagate saliency from object level down the hierarchy to parts by
 incorporating again a factor proportional to the size of the children and
 accumulating it downwards according to
 \begin{eqnarray}
  \mathcal{S}_{\text{hier}}(P) &=& \mathcal{S}_{\text{hier}}\left(\operatorname{parent}\left(P\right)\right) \cdot \frac{A_P}{A_P + A_Q} \nonumber\\
  &+& \mathcal{S}_{\text{peri}}(P)_Q .
 \end{eqnarray}
 Here, the whole image (the root node) is initialized with no saliency at all,
 $\mathcal{S}_{\text{hier}}(\text{root}) = 0$.
 This hierarchical saliency scheme can be calculated in a single run over the
 binary partition tree and is very cheap to compute.
 
\subsection{Maximum Saliency Projection to a Planar Map}
\label{sec:max_sal_map}
 Looking at a single pixel, it belongs to a certain path $\pi$ from the corresponding
 leaf segment to the root of the binary partition tree. We define its saliency to be the
 maximum of hierarchical saliency along that path. This yields a kind of min-max
 optimization, since the tree was built in a way which minimized contrast between
 merged nodes. The maximum saliency can be propagated down to the leaves in a
 single run through the tree and thus rendered into a single planar saliency map
 very quickly in linear time:
 \begin{equation}
  \mathcal{S}_{\text{max}}(P) = P\in\pi: \max_{p_i\in\pi}\left(\mathcal{S}_{\text{hier}}\left(p_i\right)\right) .
 \end{equation}
 An important property of our planar saliency is that it is deduced from the most
 salient object or part a pixel belongs to. Consequently, other pixels belonging
 to the same object and parts share exactly the same saliency value and the borders
 are crisp.\\
 Furthermore, one could
 easily deduce information
 about the extent of local structures: for each pixel, what is the most salient
 segment it belongs to? Please note that this attribution to pixels is more
 information than a planar saliency map is able to represent, since we deal with
 nested segmentations. It is a unique feature of our hierarchical saliency to
 provide a saliency decomposition of the image. In this paper, we demonstrate how to
 employ it for assisted object grasping with a single click object segmentation.

\section{Saliency Assisted Object Grasping}
Assisted teleoperation can support the operator in several ways, e.g. through object segmentation, trajectory generation, collision avoidance, or goal precision. For this work only supervisory control of the robot's mostly autonomous manipulation and segmentation capabilities is needed. We leverage the operators high-level cognitive capabilities without the disturbance of caring about low-level control.

\subsection{Graphical User Interface}
For visualizing the scene and enabling the operator to control the grasping task, we implemented a point-and-click interface that is easy to use. It runs remotely on a separate computer as a ROS based rqt plugin. Therefore, all communication between the interface and the segmentation and grasping subroutines is realized via ROS topics and messages. To reduce communication load of the network and simplify usage for the operator the interface works solely on color images of the scene. The grasping subroutine also needs depth information, e.g. from a RGB-D camera, but uses the sensor input only locally on the robot. As has previously been shown in literature, such an interface operating on 2D visualization is well suited for assisted teleoperation of a manipulator~\cite{leeper2012strategies}, \cite{you2012assisted}.

\begin{figure}
  \begin{center}
   \includegraphics[width=0.7\linewidth]{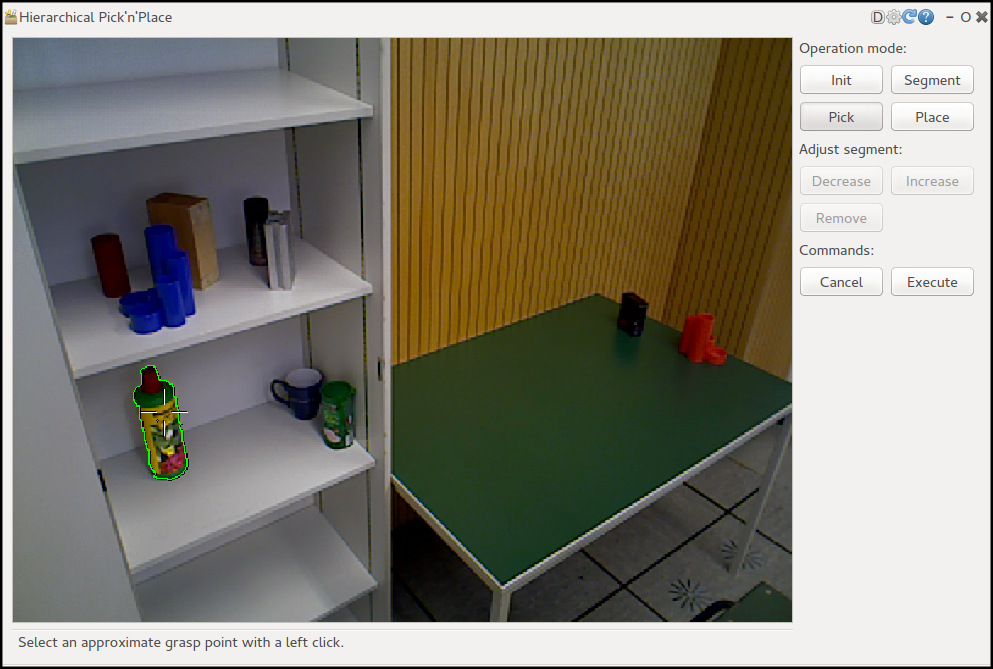}
  \end{center}
  \caption{Screenshot of the graphical user interface with a green border marking the active segment, the bottle in the lower cabinet section.}
  \label{fig:gui}
 \end{figure}

As seen by the operator, the interface itself consists mostly of a display of the scene. Depending on the currently selected operation mode the shown image is either taken from the video live stream or a rendered still image from the segmentation routine. For interaction the operator performs single mouse clicks directly on the shown image. Additionally needed commands, like canceling the current segmentation, are issued either through keyboard shortcuts or buttons. The design decision to make keyboard control optional enables to use the interface also on a touch-only device like a tablet or in combination with some sort of head tracker as motion-impaired user. A screenshot of the interface is shown in Figure~\ref{fig:gui}.

In the following, we will give a brief overview of the workflow to grasp an unknown object and afterwards place it at a target location. Starting with a live view of the scene, a segmentation is initialized with the most salient segment corresponding to the coordinates of the first click. All possible adjustments to this segmentation will be shown rendered on the current snapshot at the time of the initial click. The operator now has the option to eventually refine the segmentation by marking additive or subtractive parts or by changing the size of the currently active segment via traversing the hierarchical segmentation tree. At any time, the current segment can also be deleted to revert false clicks. Once the operator is satisfied with the shown segmentation, an approximate grasp point on the segmented object has to be chosen and confirmed. The robot will now autonomously try to grasp the segmented object while the operator can supervise the execution through a live view of the scene. After a successful grasp attempt the operator can specify a target position for placing the object. Execution of the place operation is done autonomously again. All possibly needed interaction is available through the interface, e.g. resetting the segmentation, republishing the current selection, or choosing a different grasp point. Further details about the underlying grasp process will be outlined in the following section.

\subsection{Grasping Technique}
The specified grasp point and object outline determined in the graphical user interface are passed to the grasp subroutine. It estimates the remaining degrees of freedom, including gripper orientation as well as directions and distances for the approach, lifting, and deploying movement. Secondly, the grasp subroutine triggers the planning and execution of corresponding manipulator motions.
The input of the grasp subroutine for object picking consists of three parts. The first is a 2-dimensional image coordinate specifying the user's choice of a preliminary grasp point, which will possibly be changed later on. The second part is a binary image mask which represents the object that is to be grasped. Lastly, there is a matching ordered point cloud of the scene.
First, we filter the pointcloud to remove calibration residues. Then, we calculate the surface normal at the given grasp point and align the gripper orientation to it. We target to place the tool-center-point approximately at the clicked grasp point, whereby the end-effector pose is already roughly defined.
Additionally, we need to set the turn orientation around that normal direction. We distinguish two cases: If the surface normal is nearly parallel to the gravity vector, the gripper approaches from the top, planning a so called top-grasp. Otherwise, we do a side-grasp. For that we align the gripper jaws perpendicular to the gravity vector. Furthermore, we determine the width of the object from all points belonging to the plane used for calculating the grasp normal and update the grasp point to be at the center between the rightmost and leftmost points.
In case of a top-grasp, we first extract all points on the plane used for the normal estimation and calculate their convex hull. The gripper turn orientation is then set to the angle, that results in the minimal extent of the points, which can be calculated easily using the convex hull. The final grasp point is determined at the center of the points corresponding to the minimal extent.
Additionally, our algorithm calculates the distance between the grasp point and the segment's lowest point in the direction of the gravity vector, which is needed for placing the object later on.
The approach trajectory is generated using a straight Cartesian path of 12cm length for the gripper along the normal orientation.
A collision-free path to the starting point of the Cartesian movement is planned with the ROS MoveIt! library.

For placing the object, our grasp subroutine needs one 3D point. It is looked up in the current, ordered 3D pointcloud using a 2D index specified by a click within the user interface. The subroutine then plans to place the  object grasp point in the same distance above ground, that was obtained using the initial segmentation for picking the object.
A collision-free trajectory to the pre-place position is calculated using MoveIt! again.
The motion for placing an object is almost a reversed pick operation. Starting with holding the object above the desired location, the manipulator then lowers the object down, before letting it go and retreating backwards along the main orientation of the gripper.

  \begin{figure*}[t]
   \begin{center}
   \begin{footnotesize}
    \begin{tabular}{p{0.08\linewidth}p{0.10\linewidth}p{0.08\linewidth}p{0.08\linewidth}p{0.08\linewidth}p{0.08\linewidth}p{0.08\linewidth}p{0.08\linewidth}p{0.08\linewidth}}
     image & gt. & PaTS & DRFI~\cite{DRFI} & ST~\cite{ST} & QCUT~\cite{QCUT} & RBD~\cite{RBD} & HS~\cite{HS} & CHM~\cite{CHM}
    \end{tabular}
                                                                        \end{footnotesize}
    \includegraphics[width=0.1\linewidth]{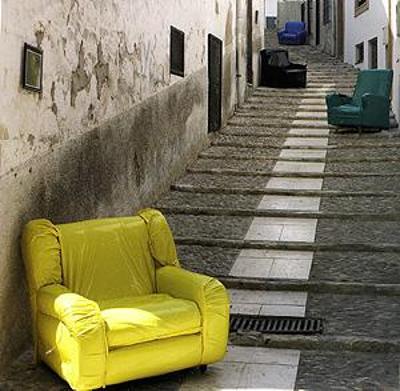}
    \includegraphics[width=0.1\linewidth]{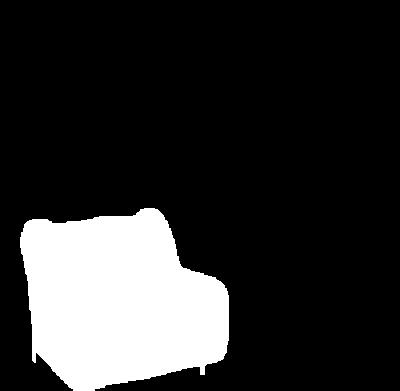}\hspace{0.02\linewidth}
    \includegraphics[width=0.1\linewidth]{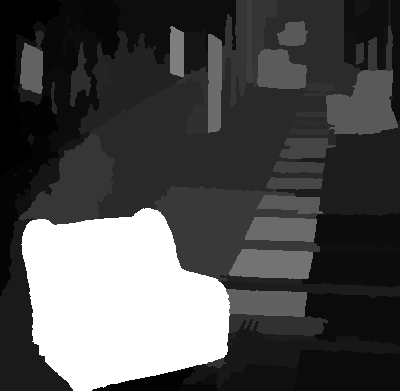}
    \includegraphics[width=0.1\linewidth]{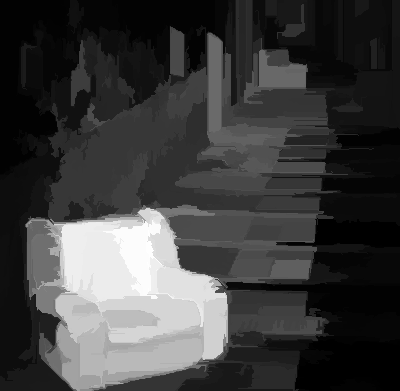}
    \includegraphics[width=0.1\linewidth]{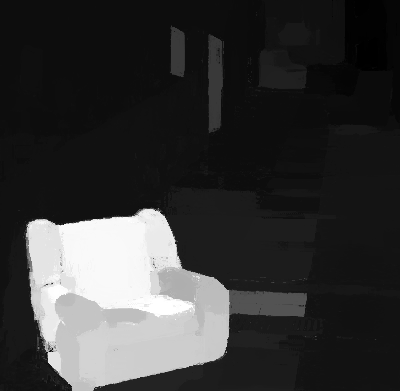}
    \includegraphics[width=0.1\linewidth]{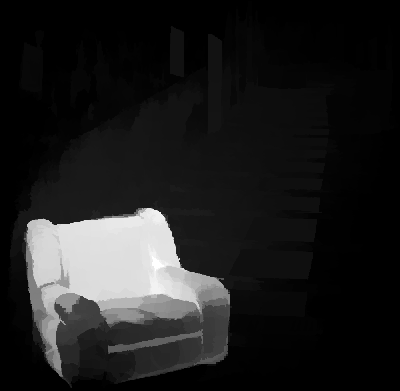}
    \includegraphics[width=0.1\linewidth]{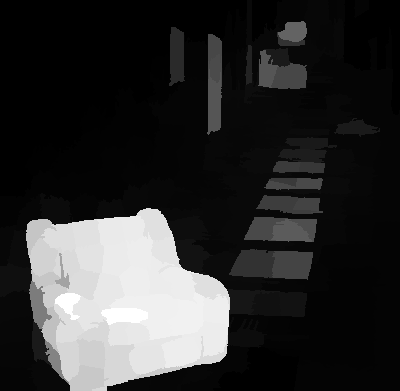}
    \includegraphics[width=0.1\linewidth]{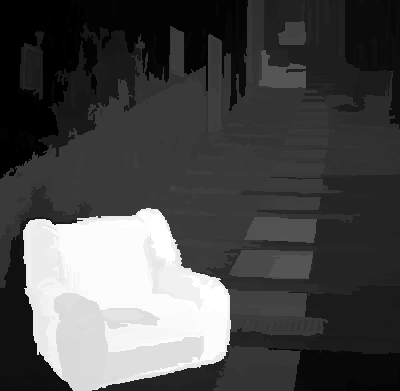}
    \includegraphics[width=0.1\linewidth]{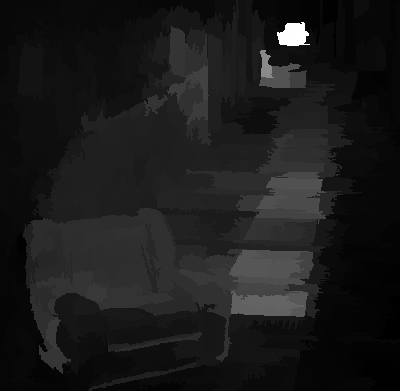}\\\vspace{1mm}
    \includegraphics[width=0.1\linewidth]{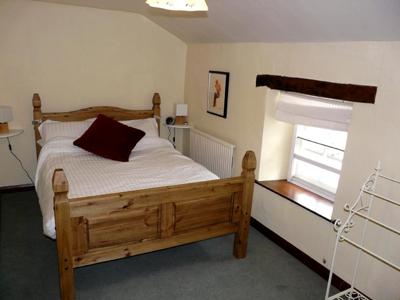}
    \includegraphics[width=0.1\linewidth]{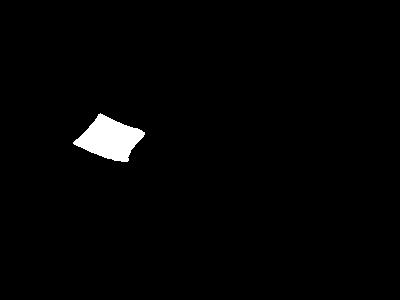}\hspace{0.02\linewidth}
    \includegraphics[width=0.1\linewidth]{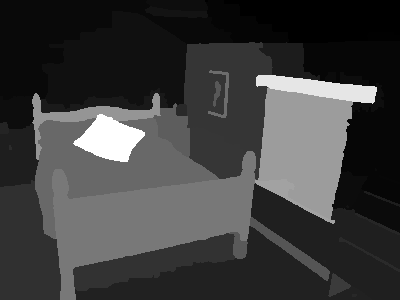}
    \includegraphics[width=0.1\linewidth]{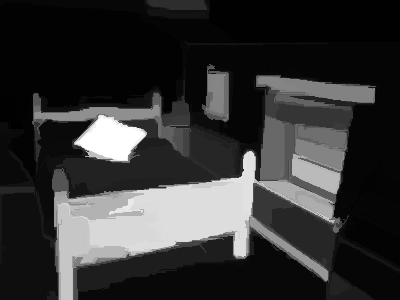}
    \includegraphics[width=0.1\linewidth]{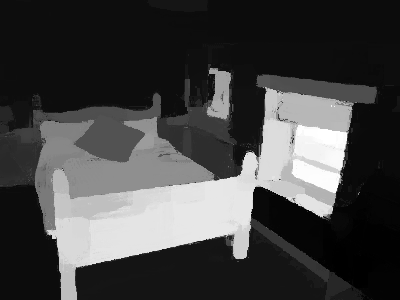}
    \includegraphics[width=0.1\linewidth]{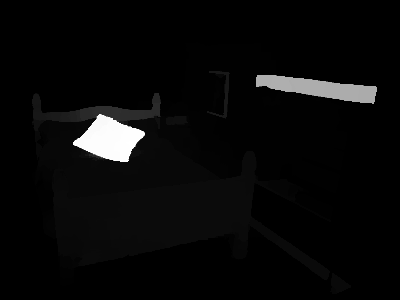}
    \includegraphics[width=0.1\linewidth]{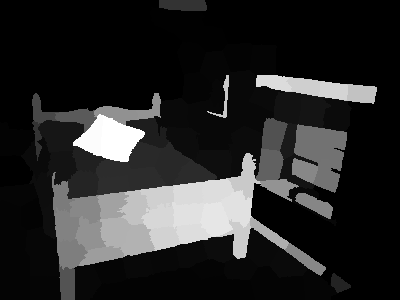}
    \includegraphics[width=0.1\linewidth]{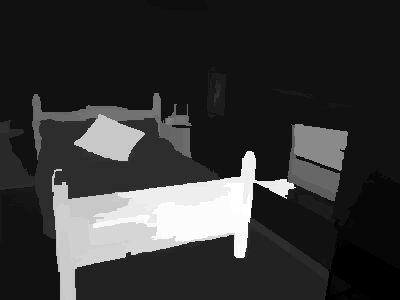}
    \includegraphics[width=0.1\linewidth]{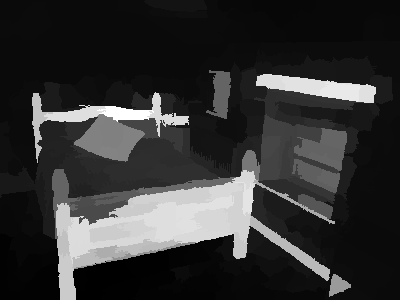}\\\vspace{1mm}
    \includegraphics[width=0.1\linewidth]{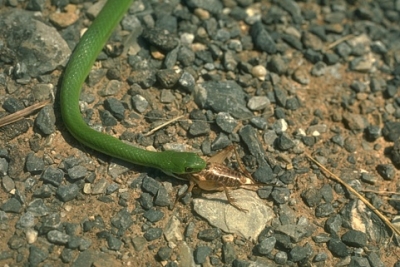}
    \includegraphics[width=0.1\linewidth]{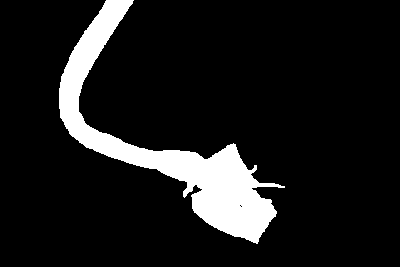}\hspace{0.02\linewidth}
    \includegraphics[width=0.1\linewidth]{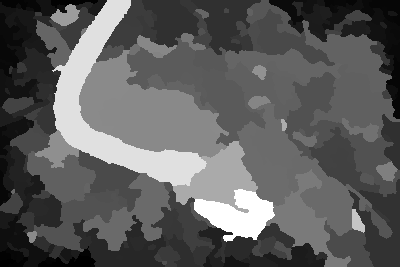}
    \includegraphics[width=0.1\linewidth]{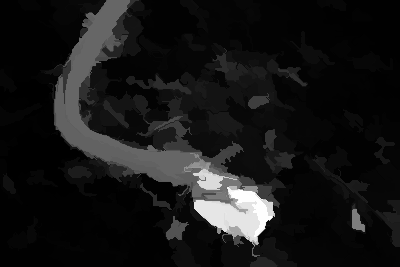}
    \includegraphics[width=0.1\linewidth]{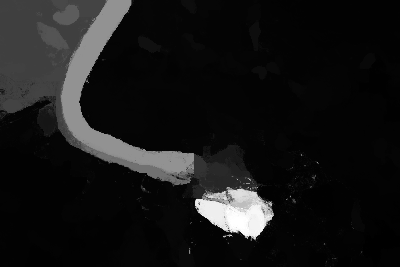}
    \includegraphics[width=0.1\linewidth]{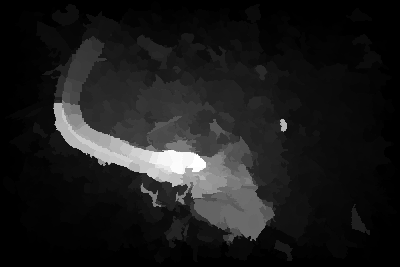}
    \includegraphics[width=0.1\linewidth]{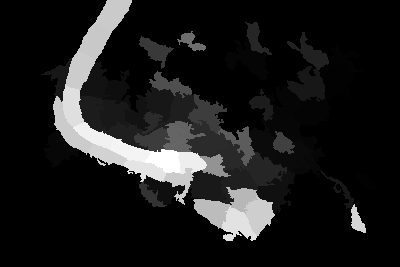}
    \includegraphics[width=0.1\linewidth]{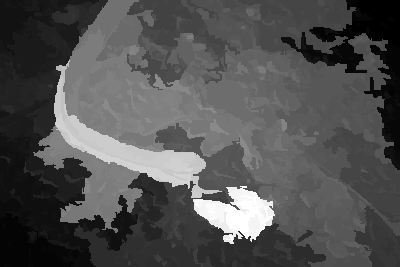}
    \includegraphics[width=0.1\linewidth]{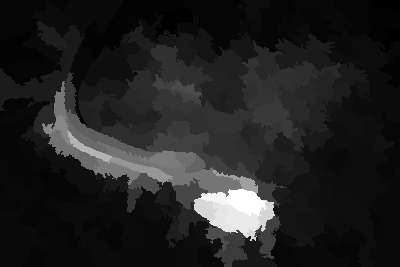}\\\vspace{1mm}
    \includegraphics[width=0.1\linewidth]{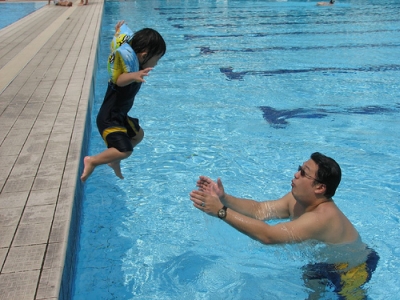}
    \includegraphics[width=0.1\linewidth]{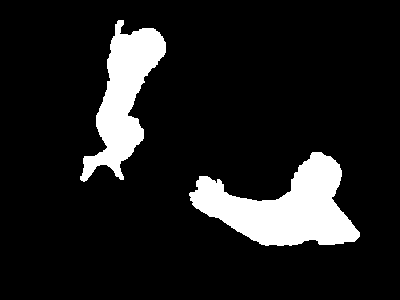}\hspace{0.02\linewidth}
    \includegraphics[width=0.1\linewidth]{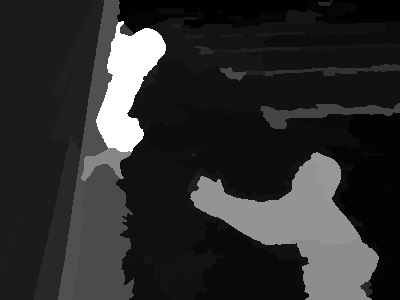}
    \includegraphics[width=0.1\linewidth]{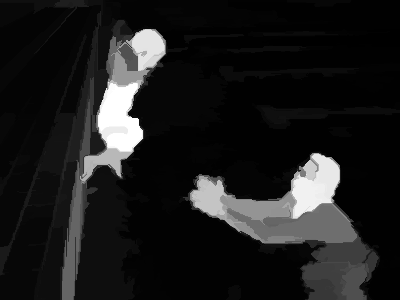}
    \includegraphics[width=0.1\linewidth]{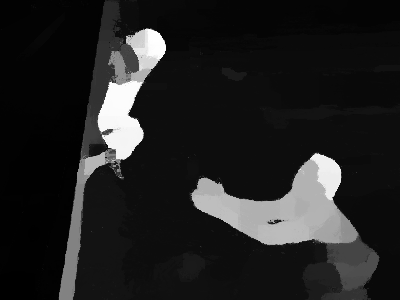}
    \includegraphics[width=0.1\linewidth]{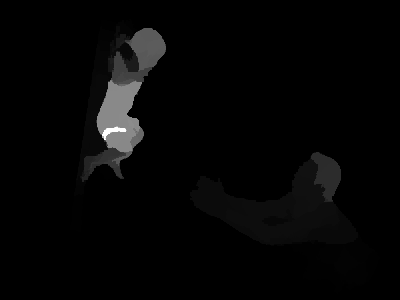}
    \includegraphics[width=0.1\linewidth]{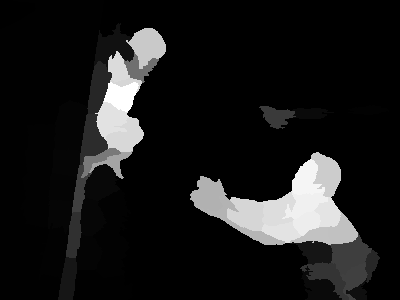}
    \includegraphics[width=0.1\linewidth]{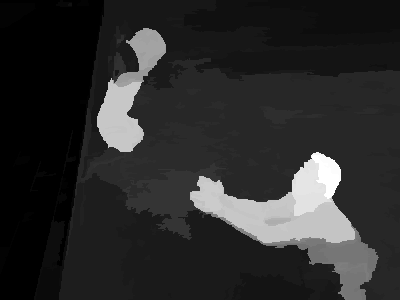}
    \includegraphics[width=0.1\linewidth]{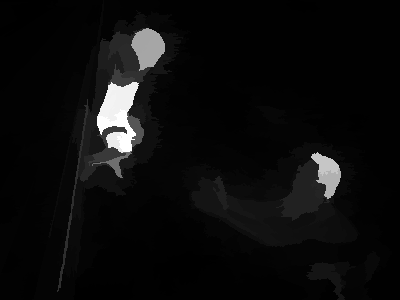}\\\vspace{1mm}
    \includegraphics[width=0.1\linewidth]{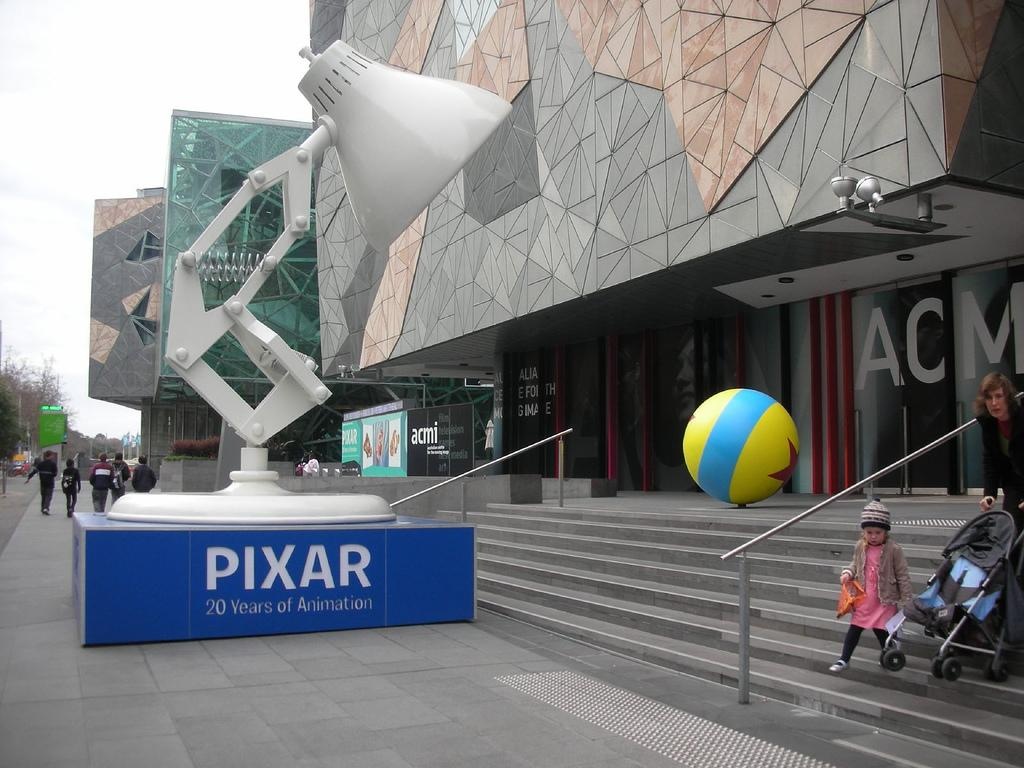}
    \includegraphics[width=0.1\linewidth]{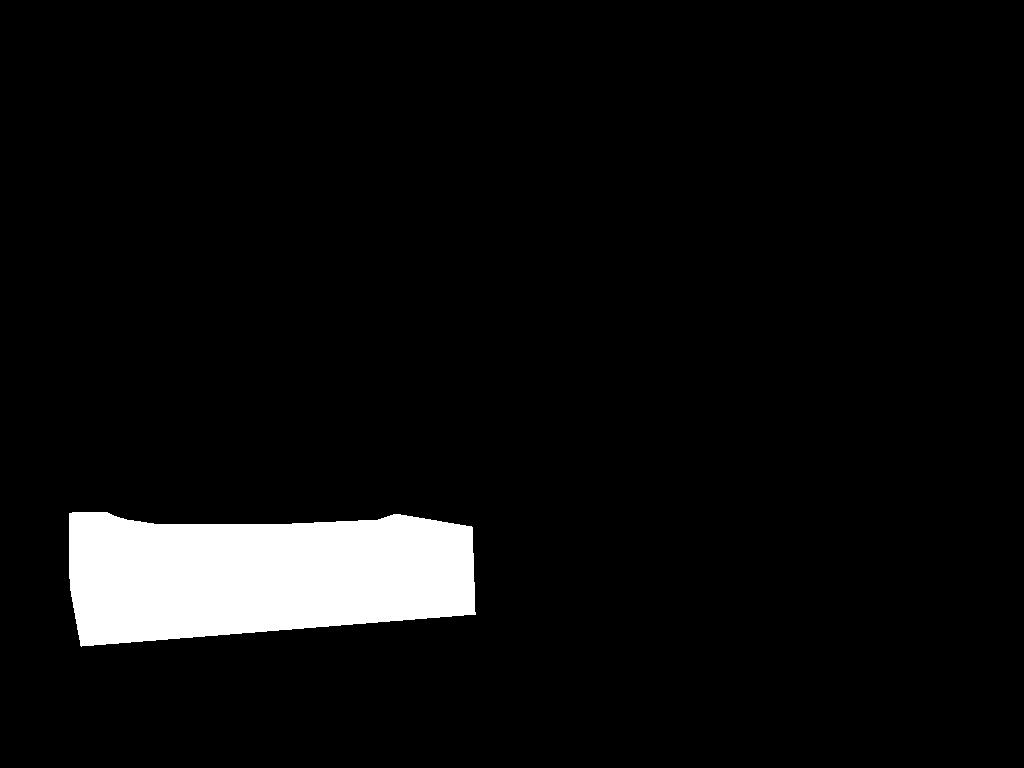}\hspace{0.02\linewidth}
    \includegraphics[width=0.1\linewidth]{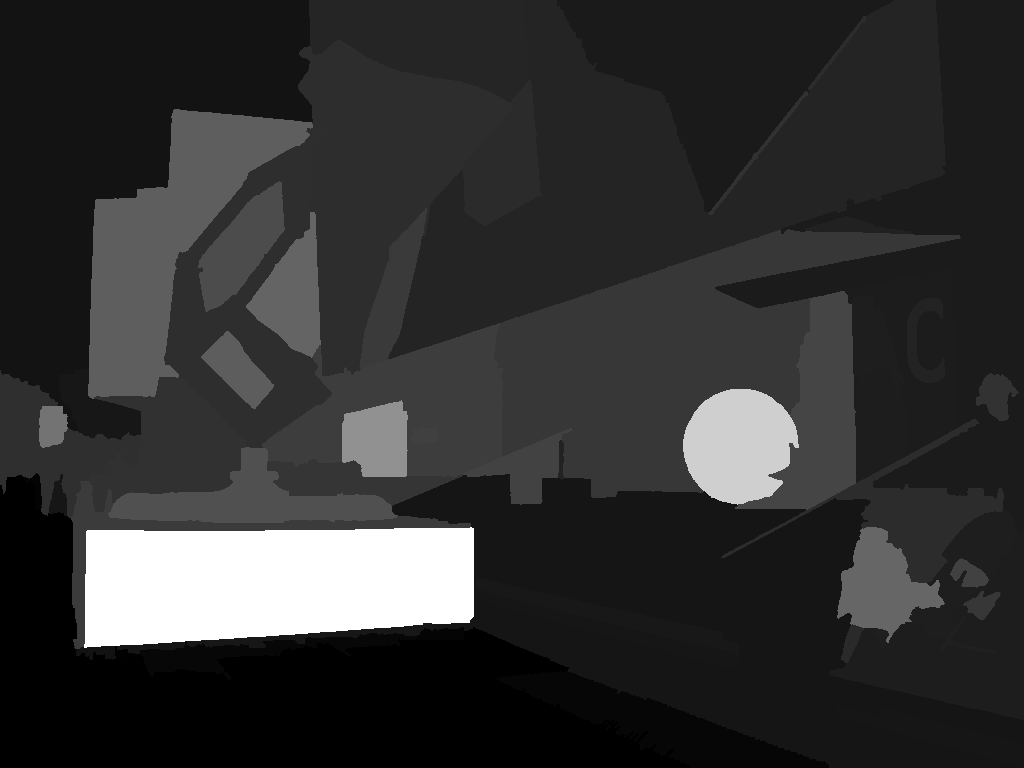}
    \includegraphics[width=0.1\linewidth]{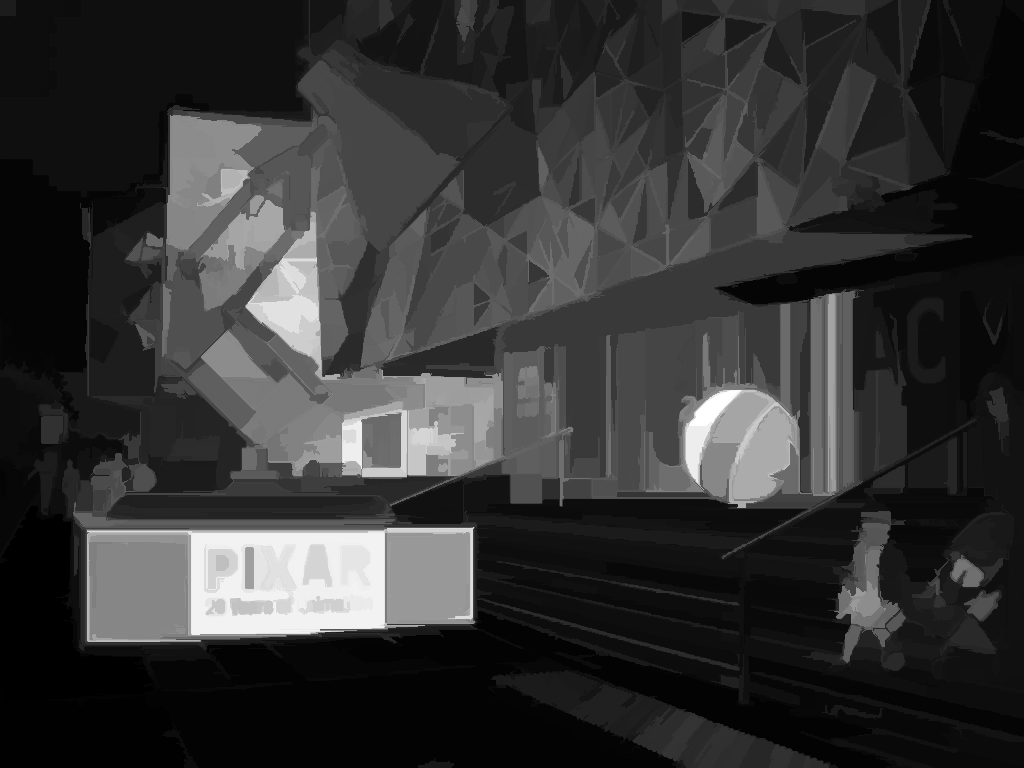}
    \includegraphics[width=0.1\linewidth]{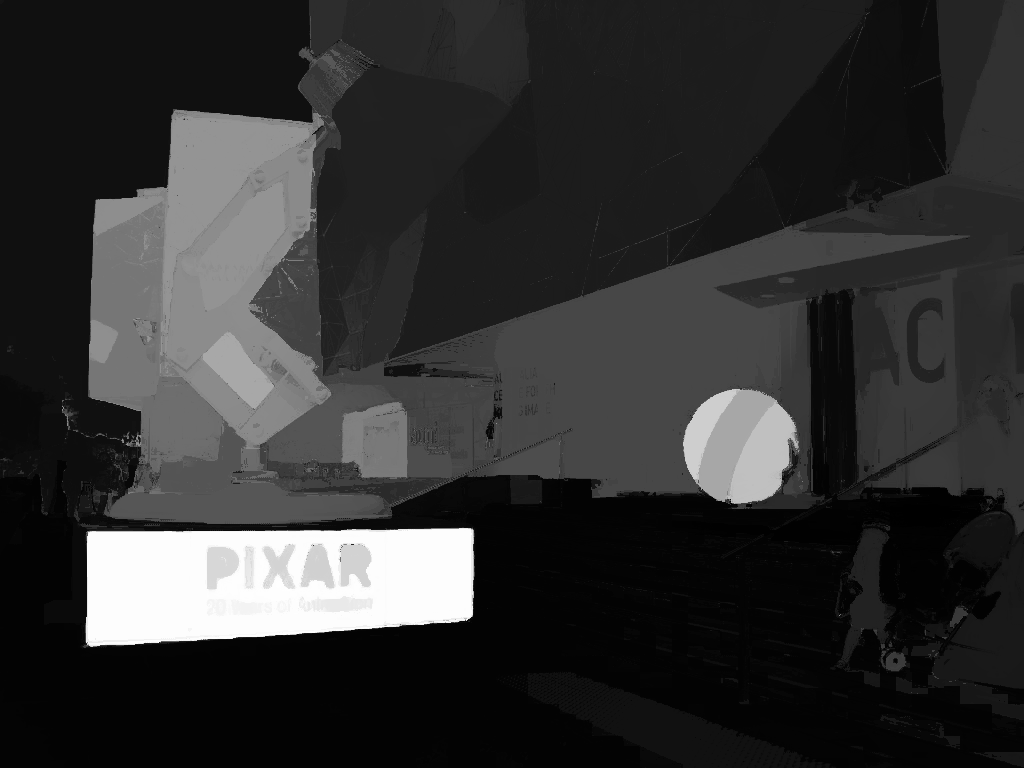}
    \includegraphics[width=0.1\linewidth]{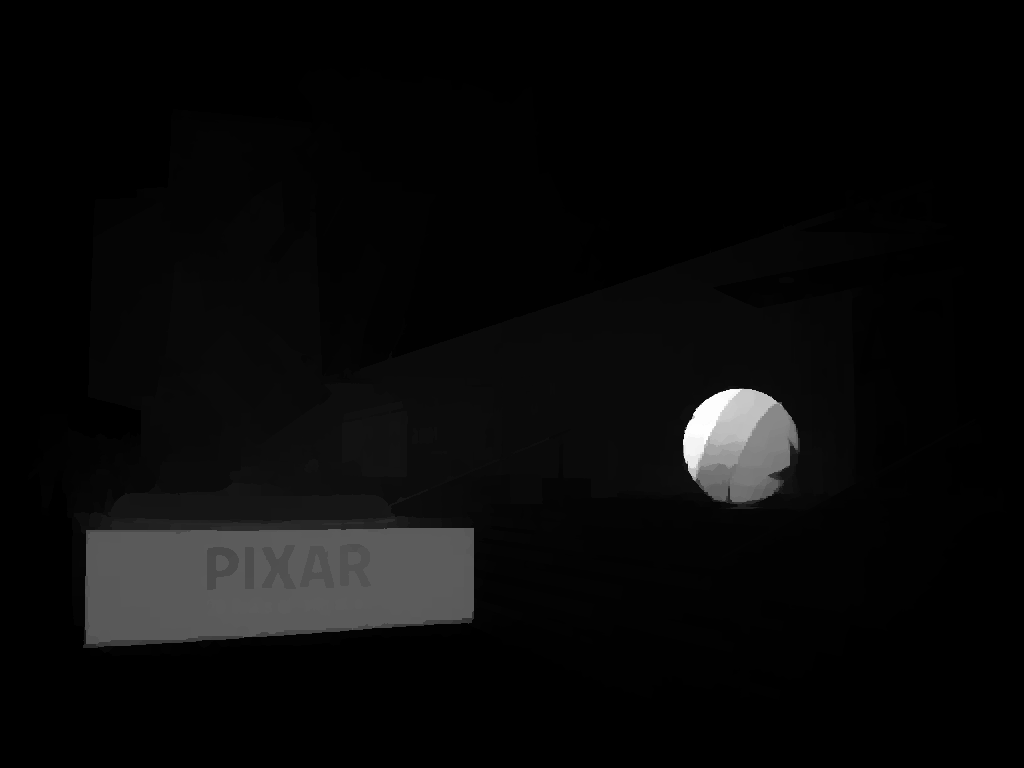}
    \includegraphics[width=0.1\linewidth]{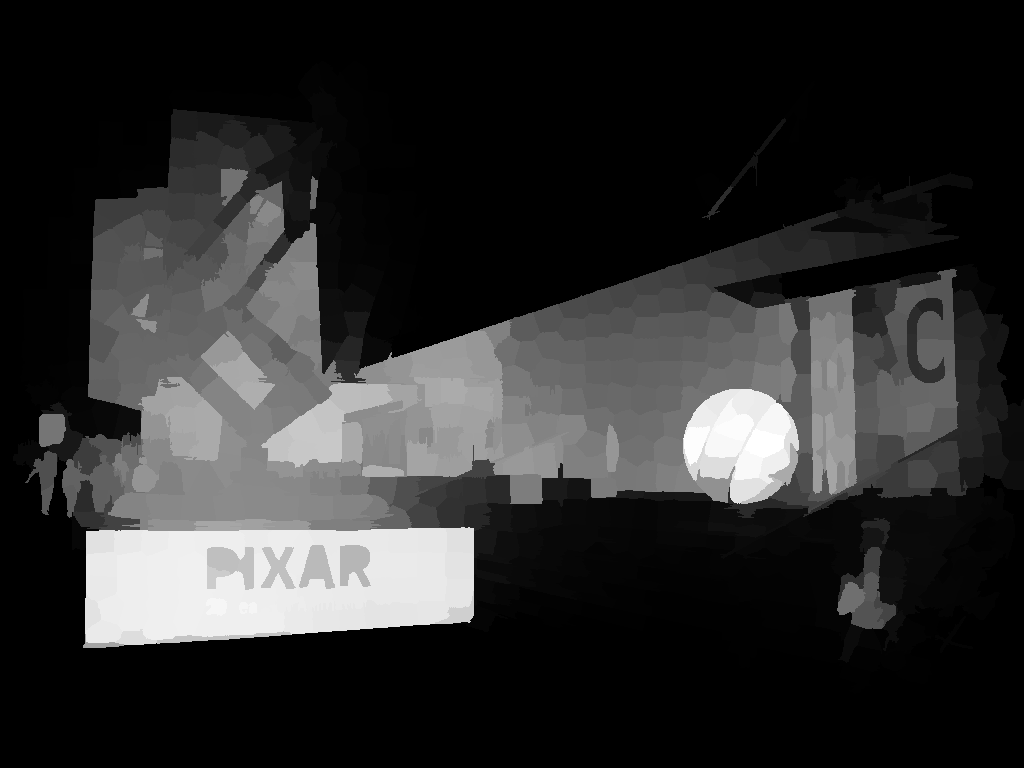}
    \includegraphics[width=0.1\linewidth]{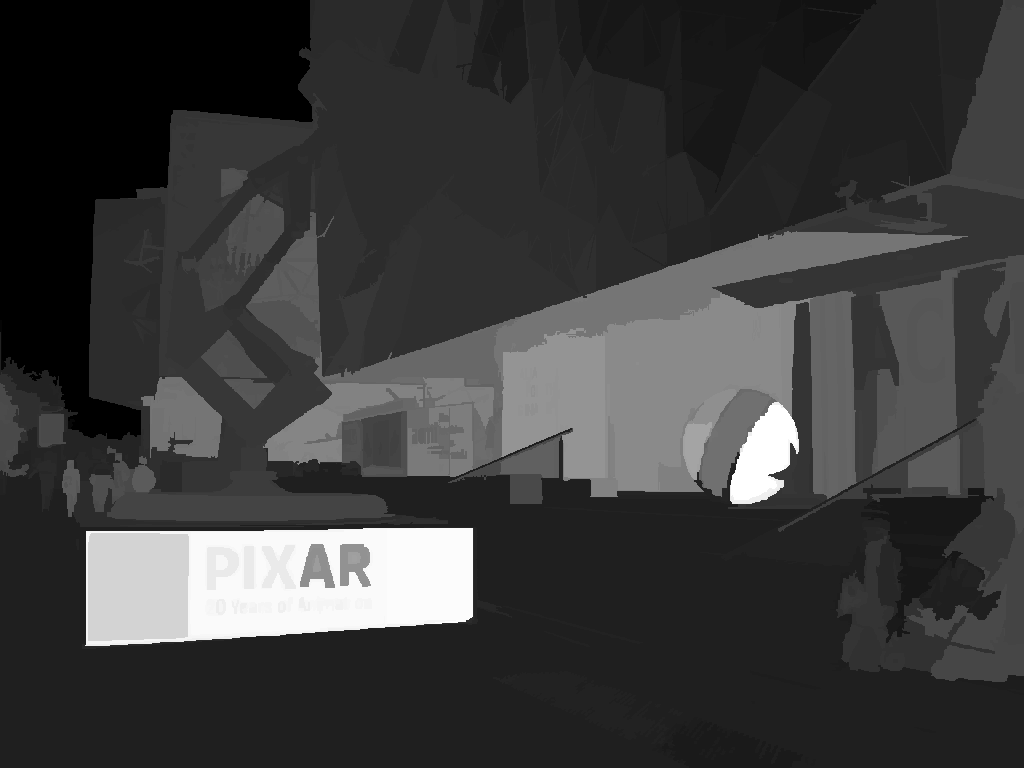}
    \includegraphics[width=0.1\linewidth]{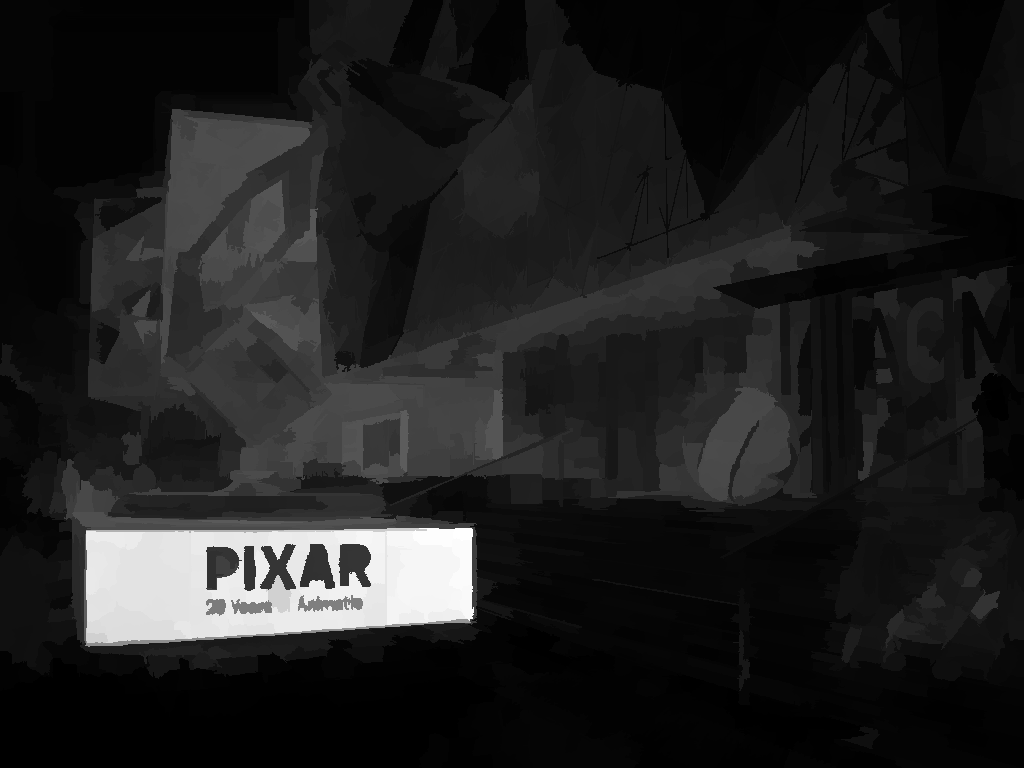}\\\vspace{1mm}
    \includegraphics[width=0.1\linewidth]{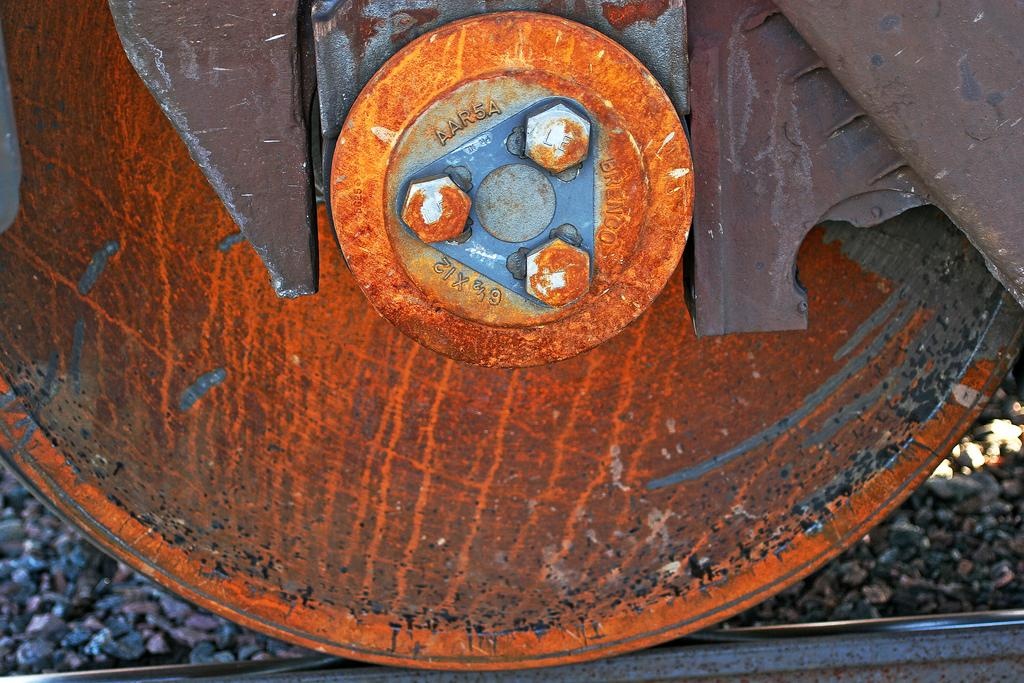}
    \includegraphics[width=0.1\linewidth]{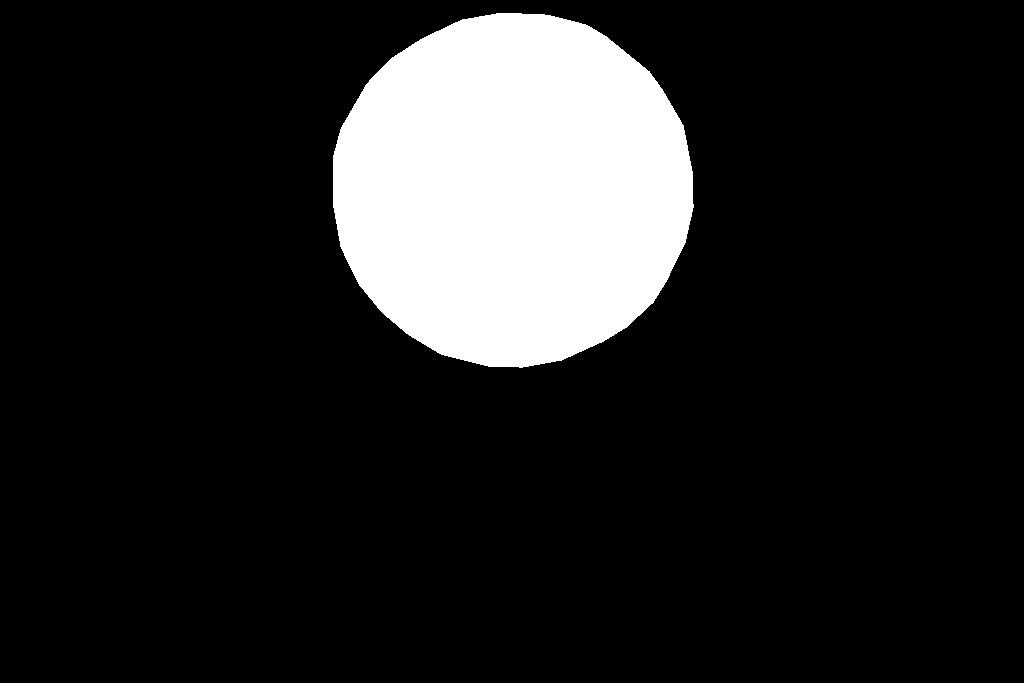}\hspace{0.02\linewidth}
    \includegraphics[width=0.1\linewidth]{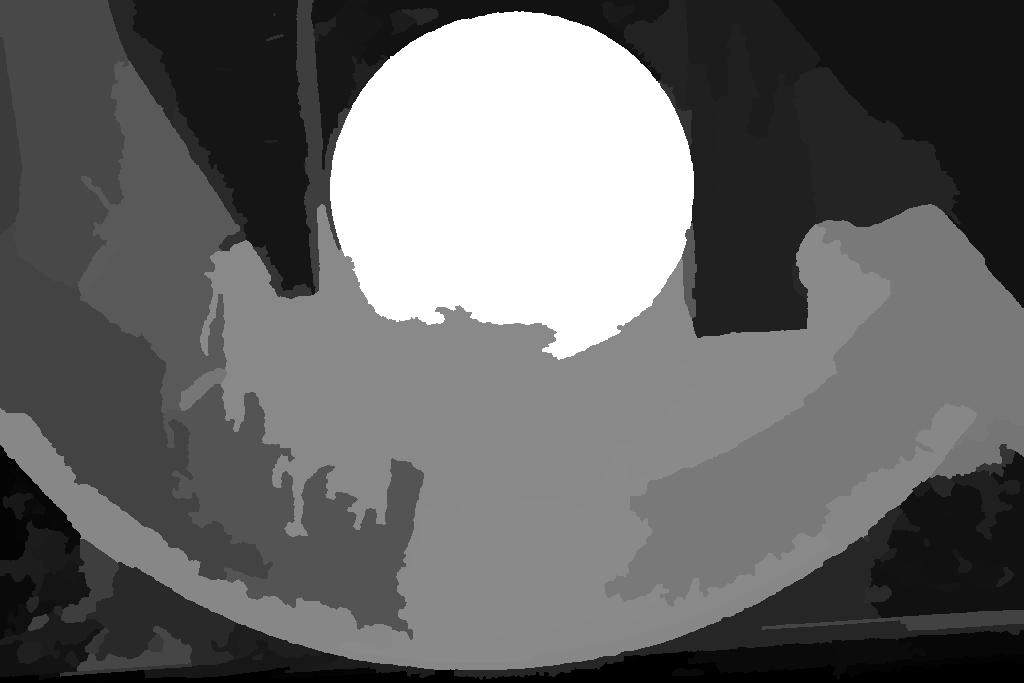}
    \includegraphics[width=0.1\linewidth]{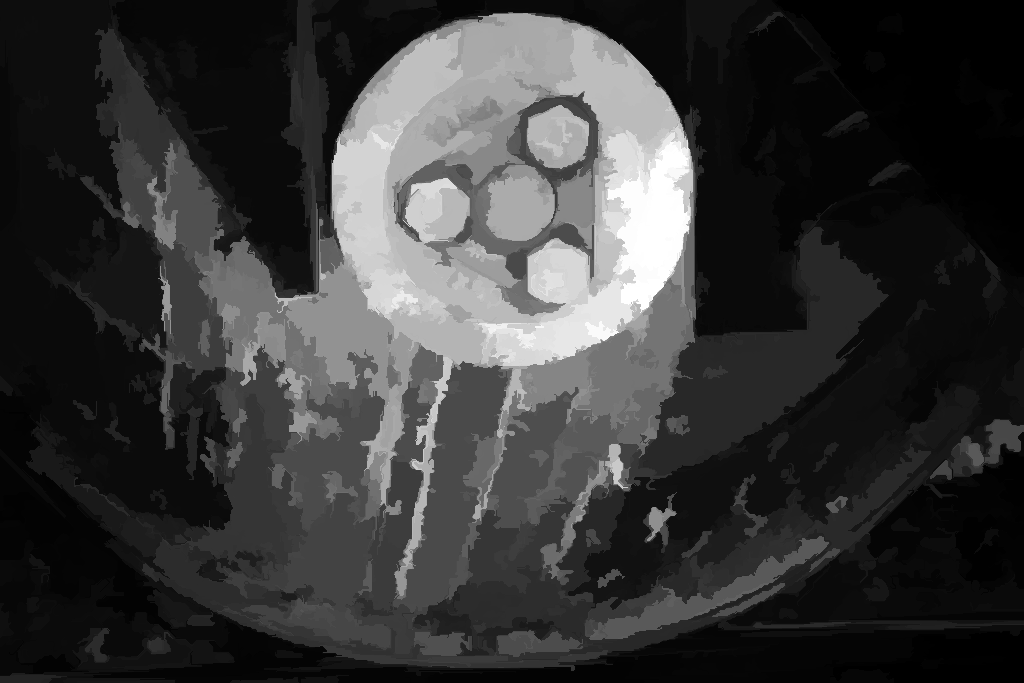}
    \includegraphics[width=0.1\linewidth]{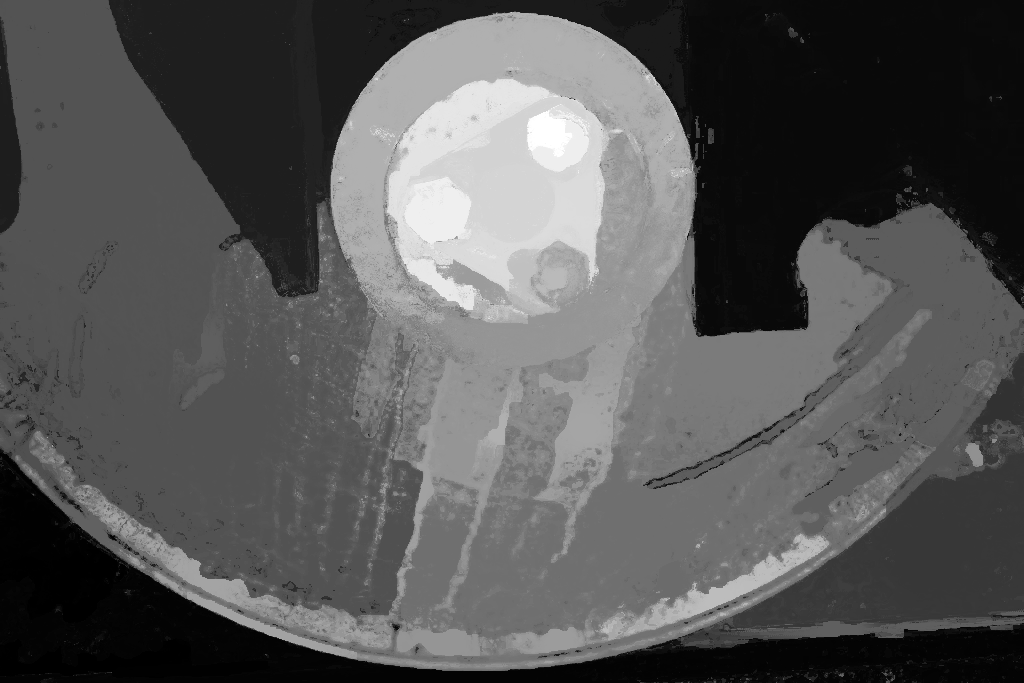}
    \includegraphics[width=0.1\linewidth]{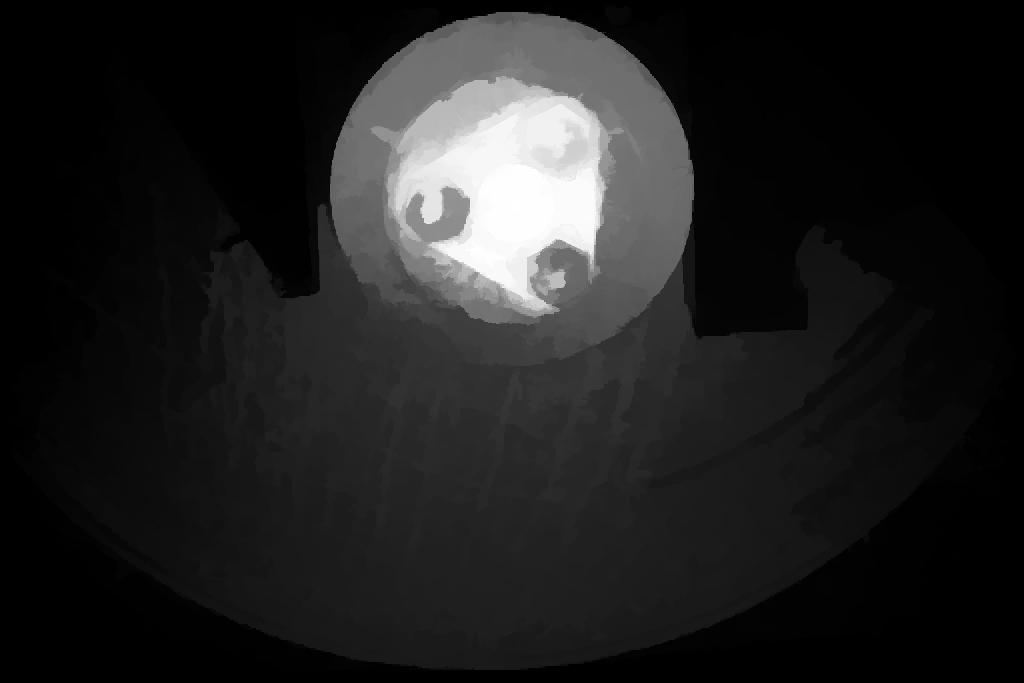}
    \includegraphics[width=0.1\linewidth]{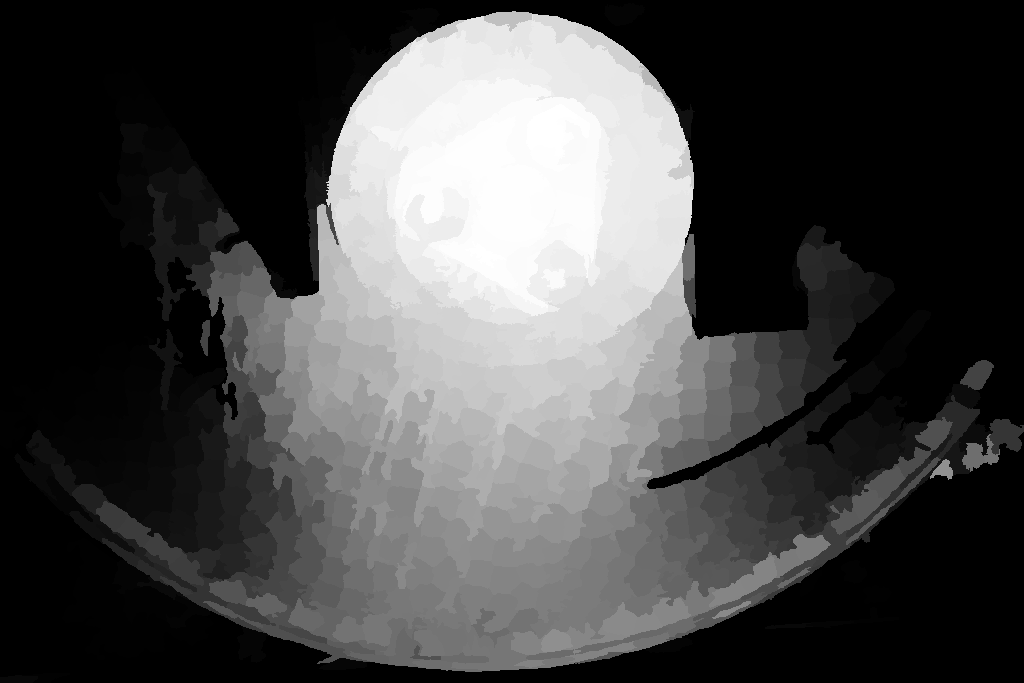}
    \includegraphics[width=0.1\linewidth]{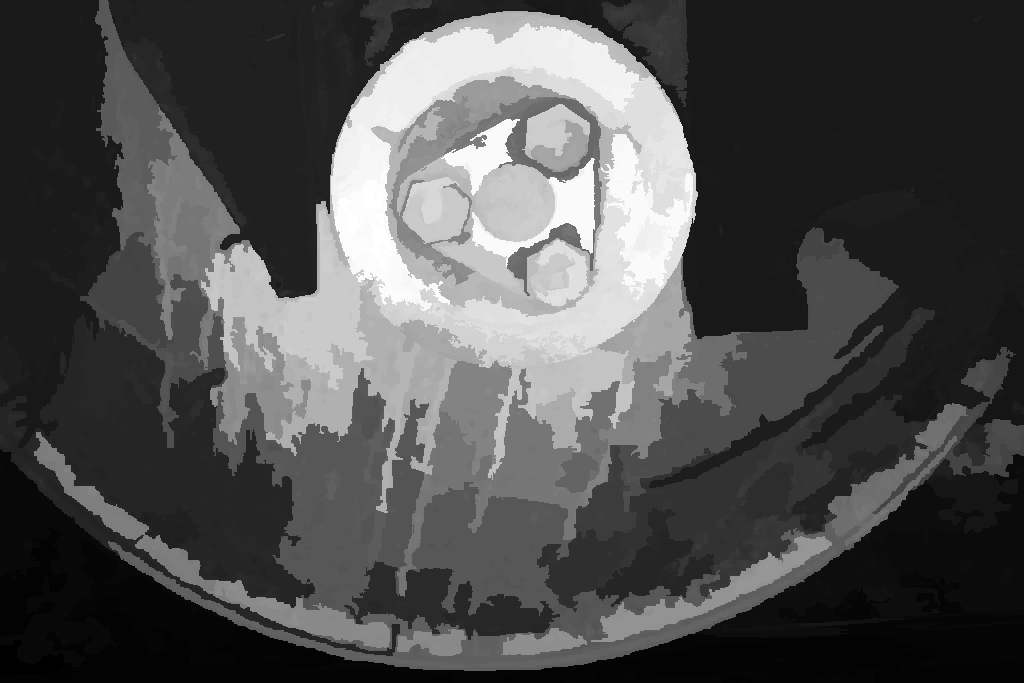}
    \includegraphics[width=0.1\linewidth]{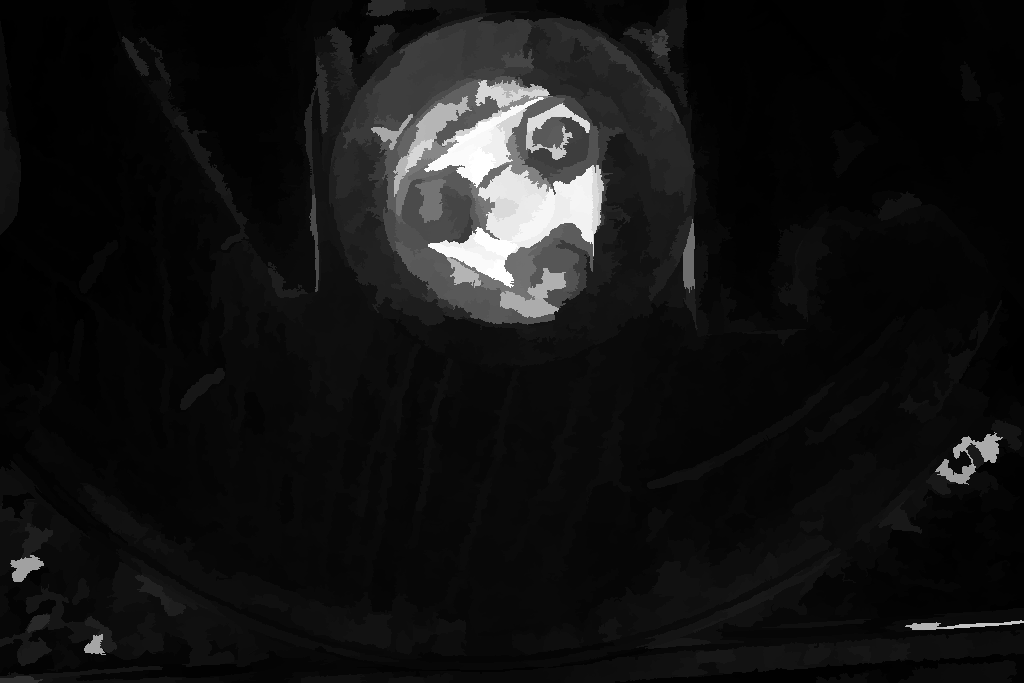}\\\vspace{1mm}
    \includegraphics[width=0.1\linewidth]{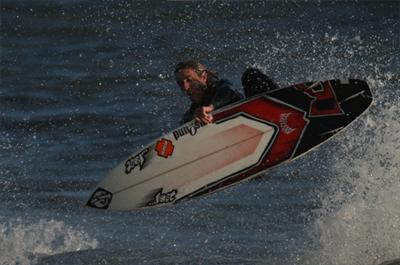}
    \includegraphics[width=0.1\linewidth]{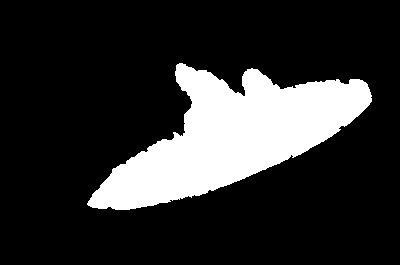}\hspace{0.02\linewidth}
    \includegraphics[width=0.1\linewidth]{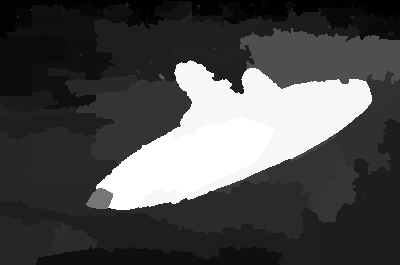}
    \includegraphics[width=0.1\linewidth]{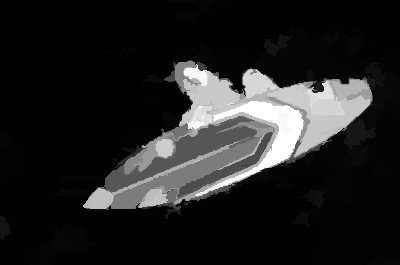}
    \includegraphics[width=0.1\linewidth]{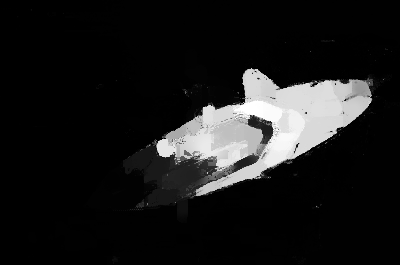}
    \includegraphics[width=0.1\linewidth]{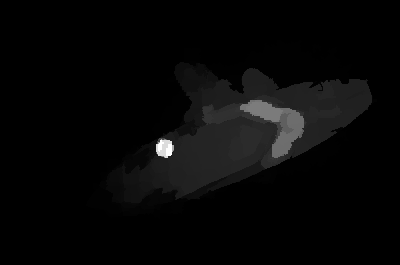}
    \includegraphics[width=0.1\linewidth]{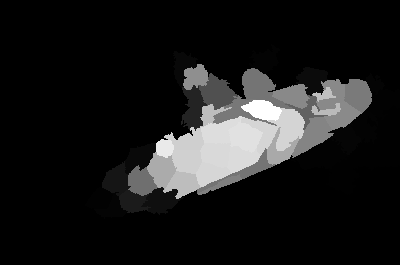}
    \includegraphics[width=0.1\linewidth]{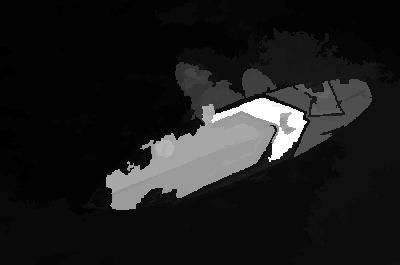}
    \includegraphics[width=0.1\linewidth]{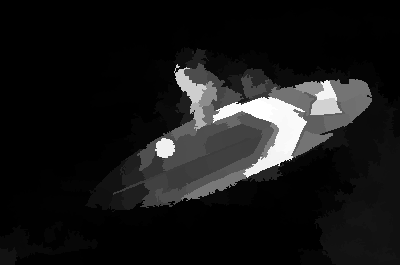}\\\vspace{1mm}
    \includegraphics[width=0.1\linewidth]{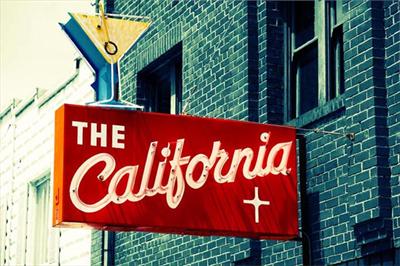}
    \includegraphics[width=0.1\linewidth]{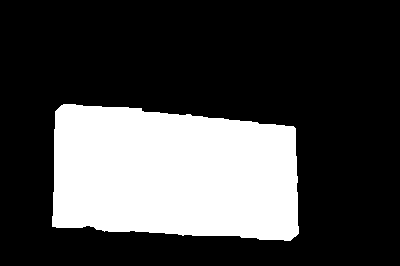}\hspace{0.02\linewidth}
    \includegraphics[width=0.1\linewidth]{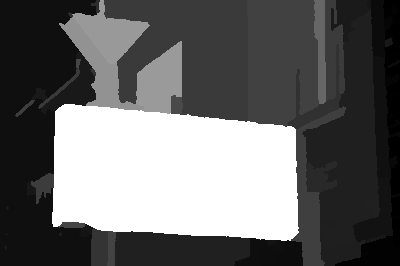}
    \includegraphics[width=0.1\linewidth]{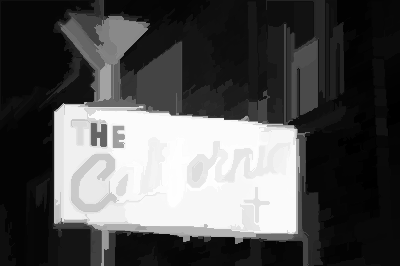}
    \includegraphics[width=0.1\linewidth]{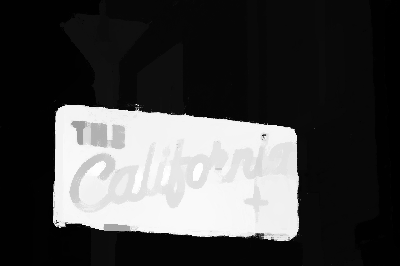}
    \includegraphics[width=0.1\linewidth]{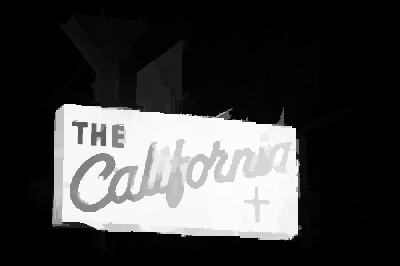}
    \includegraphics[width=0.1\linewidth]{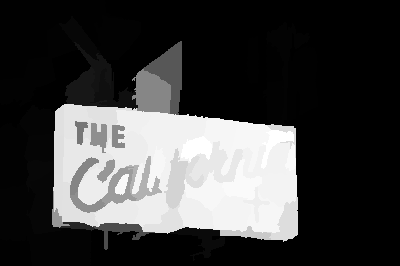}
    \includegraphics[width=0.1\linewidth]{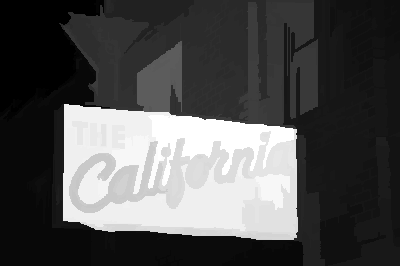}
    \includegraphics[width=0.1\linewidth]{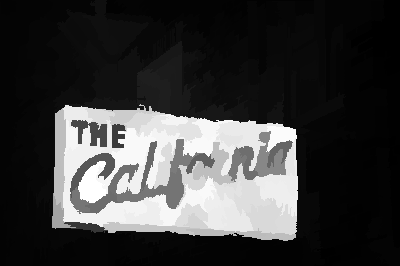}\\\vspace{1mm}
    \includegraphics[width=0.1\linewidth]{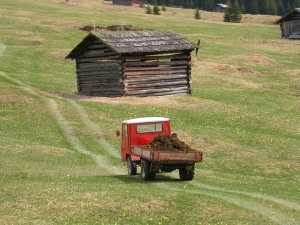}
    \includegraphics[width=0.1\linewidth]{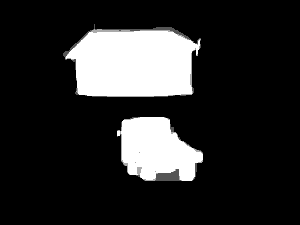}\hspace{0.02\linewidth}
    \includegraphics[width=0.1\linewidth]{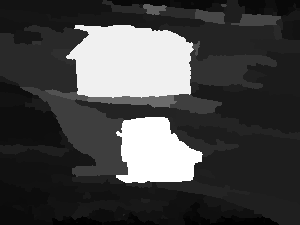}
    \includegraphics[width=0.1\linewidth]{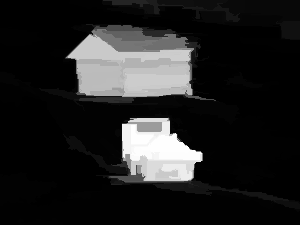}
    \includegraphics[width=0.1\linewidth]{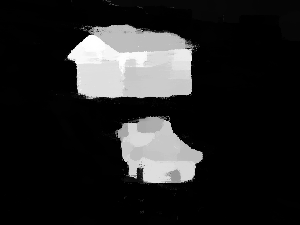}
    \includegraphics[width=0.1\linewidth]{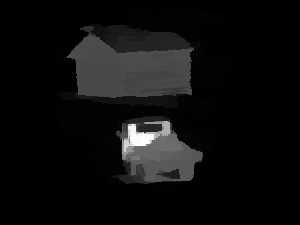}
    \includegraphics[width=0.1\linewidth]{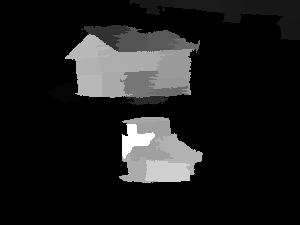}
    \includegraphics[width=0.1\linewidth]{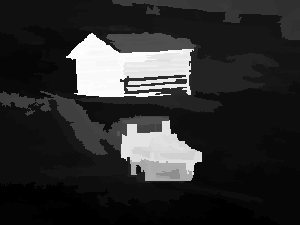}
    \includegraphics[width=0.1\linewidth]{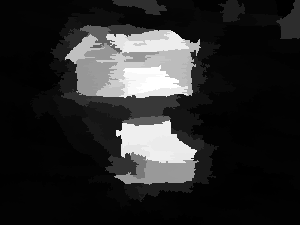}\\\vspace{1mm}
    \includegraphics[width=0.1\linewidth]{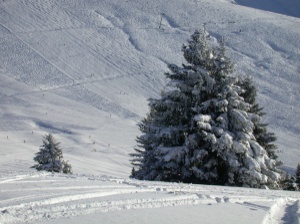}
    \includegraphics[width=0.1\linewidth]{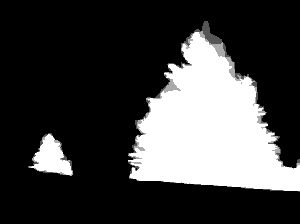}\hspace{0.02\linewidth}
    \includegraphics[width=0.1\linewidth]{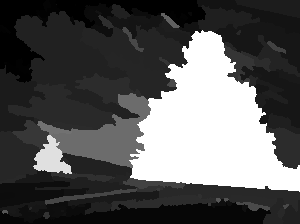}
    \includegraphics[width=0.1\linewidth]{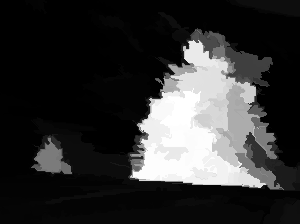}
    \includegraphics[width=0.1\linewidth]{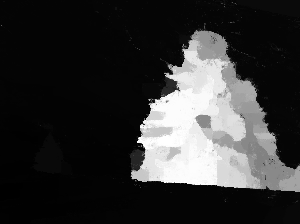}
    \includegraphics[width=0.1\linewidth]{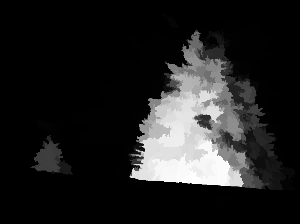}
    \includegraphics[width=0.1\linewidth]{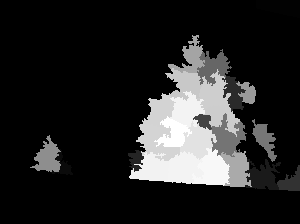}
    \includegraphics[width=0.1\linewidth]{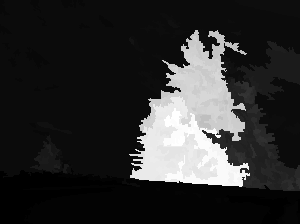}
    \includegraphics[width=0.1\linewidth]{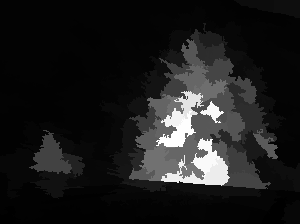}\vspace*{-3mm}
   \end{center}
   \caption{Exemplary results (saliency maps) of the compared methods. Always two images are taken top to bottom from DUT-OMRON, ECSSD, JuddDB, MSRA10K, and SED2 datasets.}
   \label{fig:result_maps}
  \end{figure*}
  
\section{Experiments and Evaluation}
 In this section, we present a twofold evaluation of our approach:
 subsection~\ref{sec:sal_eval} presents a thorough analysis of PaTS in a salient
 object detection and segmentation setting. In the following subsection~\ref{subsec:grasp_eval}, we demonstrate
 our entire system's capability for assisted object grasping. Therefore, we performed
 a number of pick and place operations and unpacked a cabinet to a desk with our mobile manipulator.

\subsection{Salient Object Detection Capability}
\label{sec:sal_eval}
 \begin{figure*}[t]
  \begin{center}
   \includegraphics[width=0.5\linewidth]{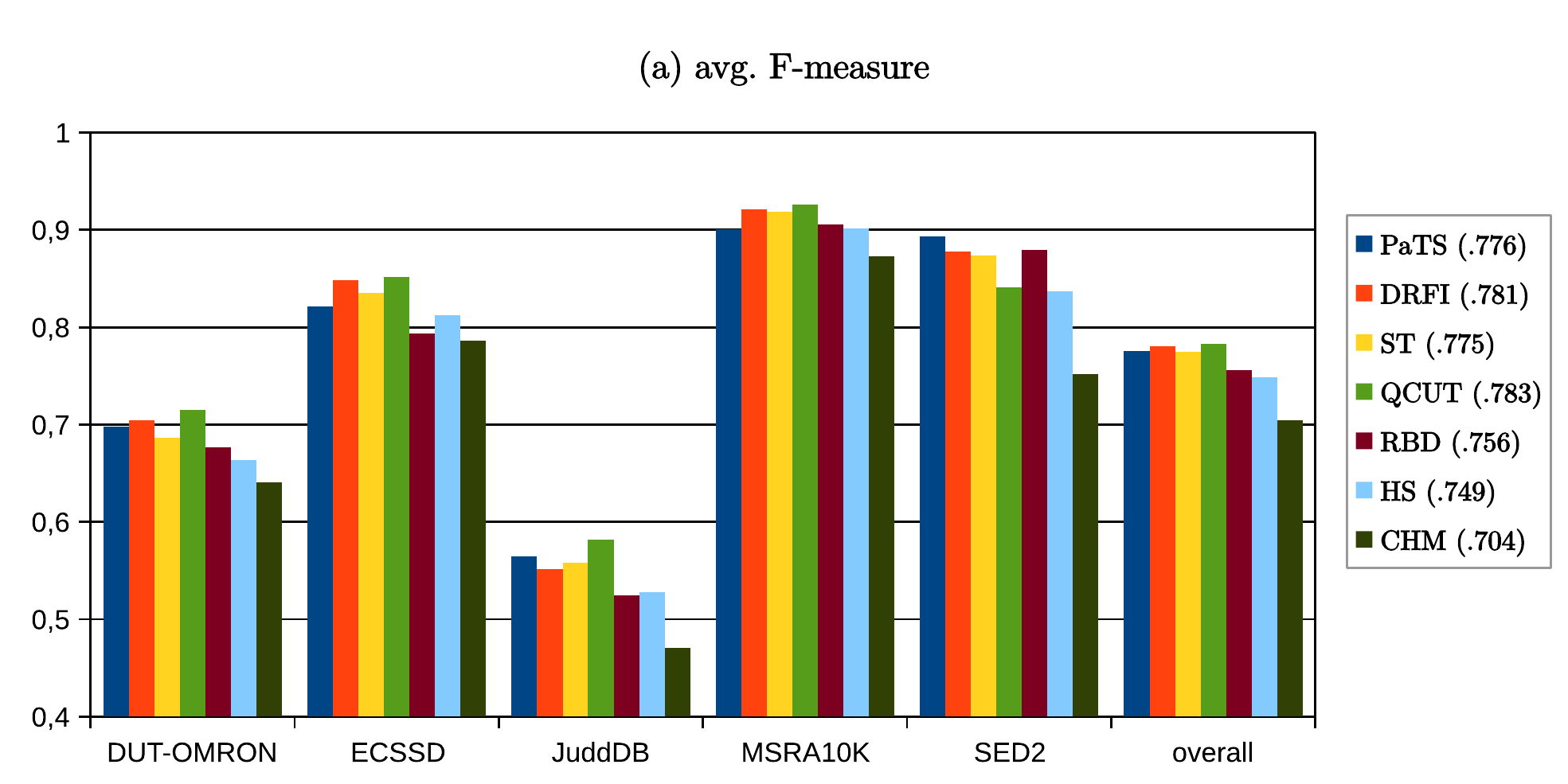}%
   \includegraphics[width=0.5\linewidth]{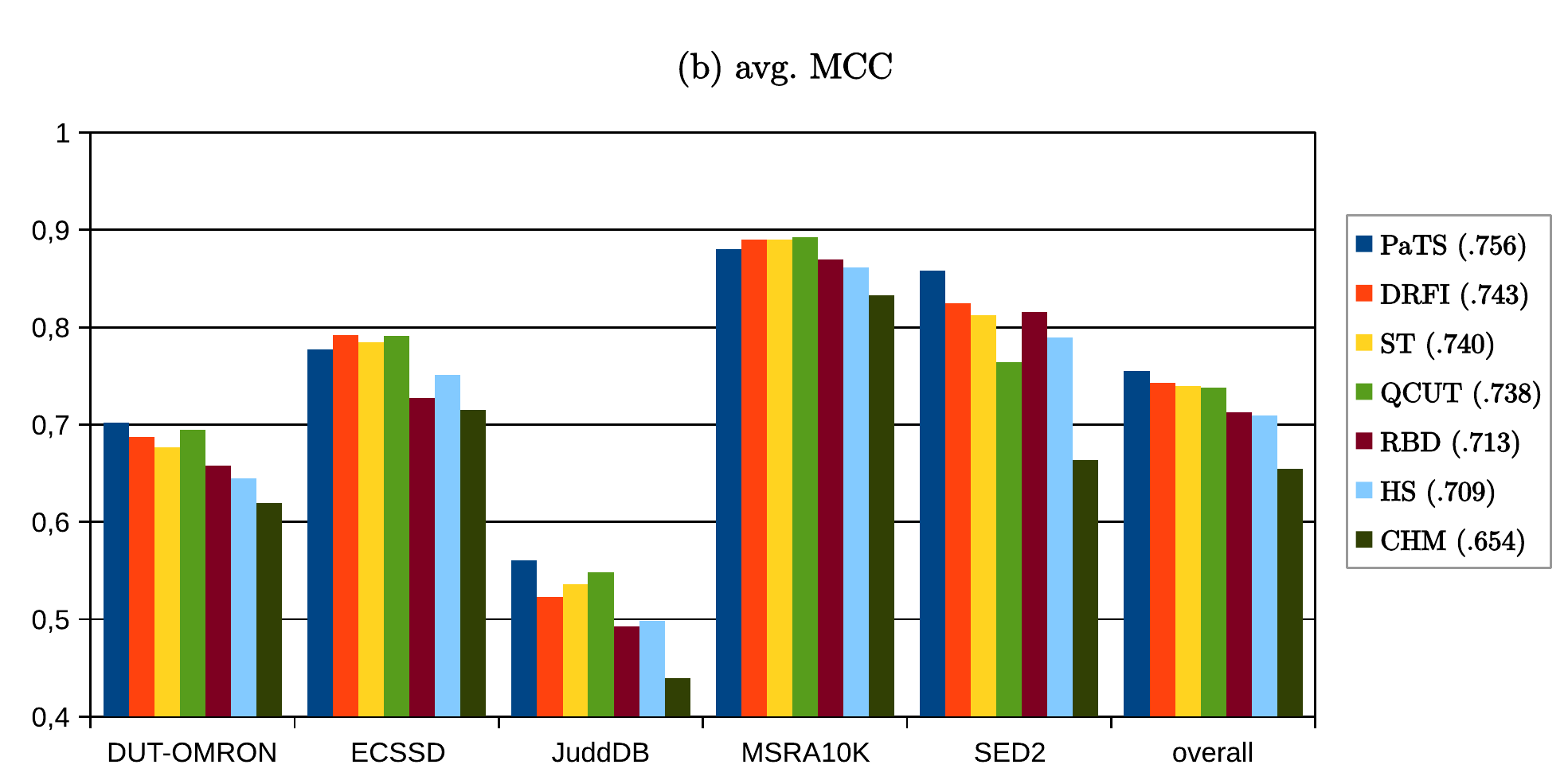}
  \end{center}
  \caption{(a) Avg. F-measure ($\beta^2=0.3$). (b) Avg. Matthews Correlation Coefficient. Both measures are computed with optimal thresholding per image.
   Values in parenthesis quantify the overall results for each particular approach.}
  \label{fig:diagram}
 \end{figure*}
 Since our PaT-Saliency is used to distinguish an object from the background in our
 application, it is appropriate to evaluate it on salient object detection and
 segmentation benchmarks. A comprehensive study in this area comparing 39 approaches
 was recently published by Borji et.~al.~\cite{SalObjBenchmark}. This work aggregated
 several previous benchmarks, namely DUT-OMRON~\cite{DUTOMRON} (5166 images), ECSSD~\cite{HS} (1000 images),
 JuddDB~\cite{JuddDB} (900 images), MSRA10K~\cite{MSRA10K} (10000 images), and SED2~\cite{SED2} (100 images).
 These images come with binary ground truth segmentations to tag which part of each
 image is
 occupied by the most salient object or the two most salient objects in some cases.
 Very valuable for following evaluations, the authors of~\cite{SalObjBenchmark} share saliency maps of all participating
 approaches on their webpage\footnote{\url{http://mmcheng.net/zh/salobjbenchmark/}}.
 We selected the best performing and most related ones for our qualitative comparison:
 CHM~\cite{CHM} (Contextual Hypergraph Modeling),
 DRFI~\cite{DRFI} (Discriminative Regional Feature Integration),
 HS~\cite{HS} (Hierarchical Saliency),
 QCUT~\cite{QCUT} (Quantum Cuts),
 RBD~\cite{RBD} (Robust Background Detection),
 and ST~\cite{ST} (Saliency Tree).
 
 Figure~\ref{fig:result_maps} shows a couple of saliency maps for visual
 comparison of the different saliency approaches.
 Our model PaTS stands out from the others in the manner that it quite successfully
 assigns an object-wide saliency value due to the maximum projection
 of the hierarchical saliency explained in section~\ref{sec:max_sal_map}.
 Furthermore, it does not only detect the most salient object, but quantifies the
 saliency of any structure in the scene, which is important for generic object
 grasping. PaTS is a simple and slick model, not incorporating specific
 assumptions on the imaging modalities. It does not even use global statistics
 of the image, but is purely based on local contrasts. Therefore, it works
 equally well on images from the web as from a robot mounted camera.

 There are several different measures how to rate the compliance between saliency
 maps and a binary ground truth. Here, we want to focus on the question how well the
 salient object could be separated from the background based on saliency.
 A critical step is how to binarize a saliency map in order to determine
 \textit{true positives}, \textit{true negatives}, \textit{false positives}, and \textit{false negatives}
 vs. the ground truth. In~\cite{SalObjBenchmark}, Borji et.~al. retained three different
 variants: best benchmark-wide fixed threshold~\cite{MSRA10K}, adaptive thresholding~\cite{achanta2009}, and
 SaliencyCut~\cite{MSRA10K}.
 The meaning of the first variant is questionable, since the saliency maps used for evaluation
 had been normalized to $\left[0\dots255\right]$ range and thus do not represent absolute values anymore.
 Furthermore, a benchmark-wide optimal threshold is artificially vague when judging the ability
 of saliency maps to separate the salient object.
 We solved both issues by using an optimal threshold per image according to the evaluation measure.
 Then, the individual image results are averaged to give the benchmark score. Finally, an overall
 score is computed as the average of all benchmark scores.
 The second and third binarization variants of~\cite{SalObjBenchmark} both define a content-adaptive way of thresholding, but thereby introduce
 additional side-effects. Likewise, using an oracle that tells the best threshold per image
 circumvents any additional dependencies on particular binarization algorithms.
 
 Widely used in statistical analysis of binary classification is the
 $\operatorname{\operatorname{F}_{\!\beta}}$-measure, which is the weighted
 harmonic mean of precision and recall. It equals
 \begin{equation}
  \operatorname{\operatorname{F}_{\!\beta}} = \frac{(1+\beta^2)\cdot \mathrm{tp}}{(1+\beta^2)\cdot \mathrm{tp} + \mathrm{fp} + \beta^2 \mathrm{fn}} .
  \label{eqn:Fb}
 \end{equation}
 The meaning of $\beta$ is that one could weight recall $\beta$ times more important than precision.
 We chose $\beta^2=0.3$ in accordance with most previous papers, which agreed
 that precision is more important than recall in this situation.
 
 Although commonly accepted, the $\operatorname{\operatorname{F}_{\!\beta}}$-measure
 has two major drawbacks: obviously, it disregards \textit{true negative} decisions,
 and it does not incorporate the size of \textit{true} and \textit{false} categories.
 In our setting, we deal with clearly smaller \textit{true} examples (the salient
 objects). Both these issues are tackled using Matthews Correlation Coefficient, which
 is formulated
 \begin{equation}
  \operatorname{MCC} = \frac{\mathrm{tp}\cdot \mathrm{tn} - \mathrm{fp}\cdot \mathrm{fn}}{\sqrt{(\mathrm{tp}+\mathrm{fp}) (\mathrm{tp}+\mathrm{fn}) (\mathrm{tn}+\mathrm{fp}) (\mathrm{tn}+\mathrm{fn})}} .
  \label{eqn:MCC}
 \end{equation}
 It returns values between $-1$ and $1$, where $0$ would be the result of random
 guesses and $1$ a perfect match.
 
 Figure~\ref{fig:diagram} depicts the quantitative outcomes for all compared
 approaches gathered on the popularly accepted datasets. They are all close together
 and achieved impressing results on an absolute scale. Still, considering the large
 number of images in the datasets, one may grade them into the leading group of
 PaTS, DRFI, ST, and QCUT, the close contenders RBD and HS, as well as the
 bottom-placed CHM. From the overall tendency, PaTS ranks third on $\operatorname{\operatorname{F}_{\!\beta}}$-measure
 and even best on MCC.
 \begin{figure}[t]
  \begin{center}
   \includegraphics[width=0.2\linewidth]{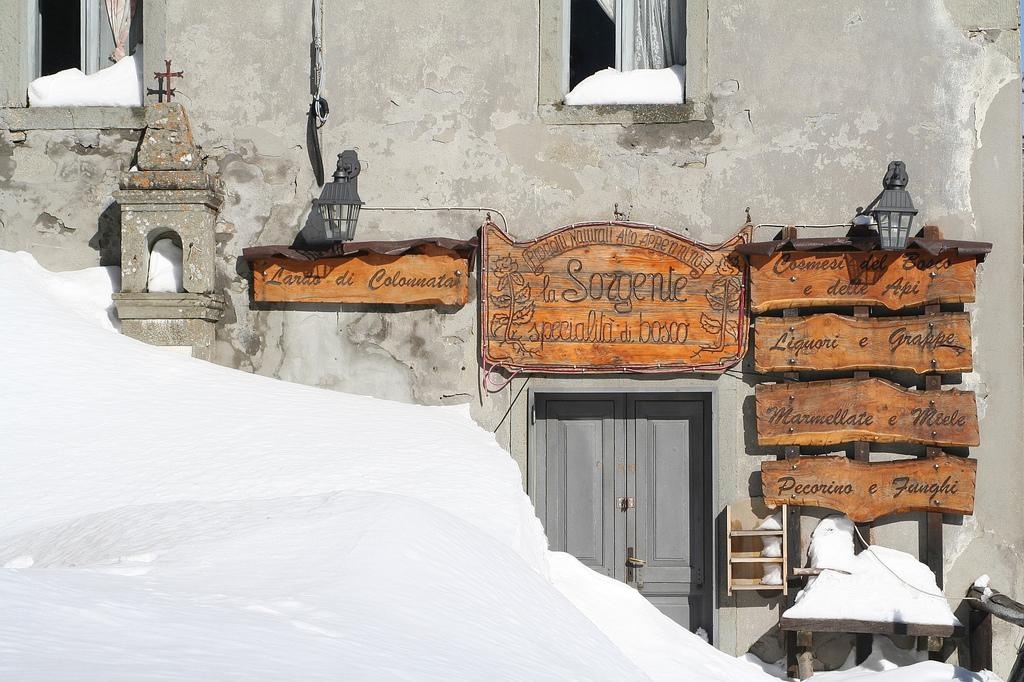}
   \includegraphics[width=0.2\linewidth]{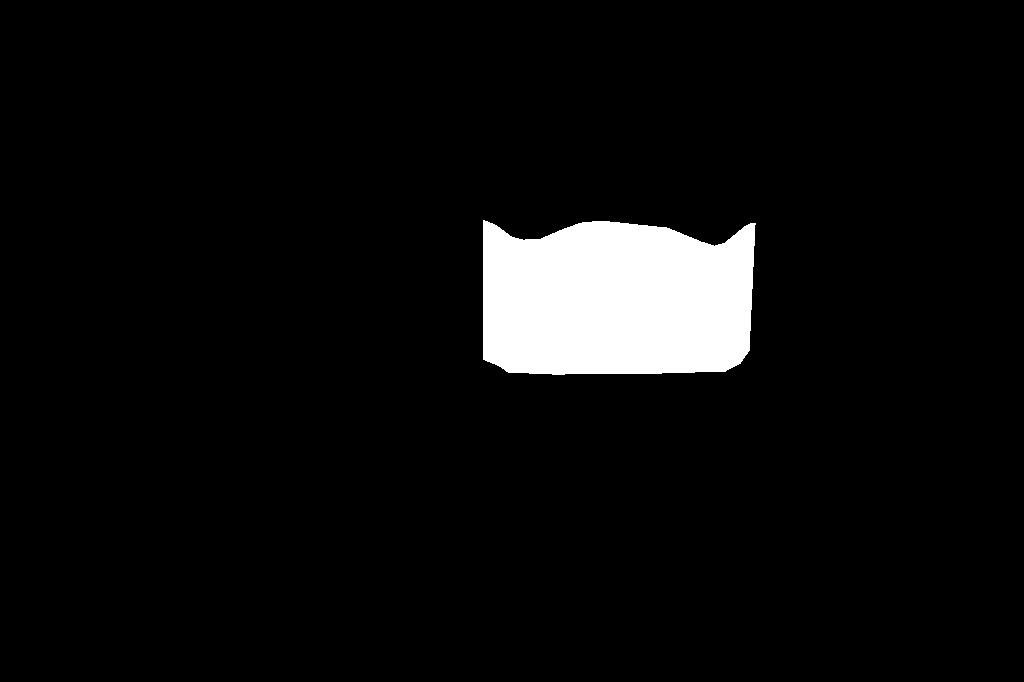}\hspace{0.02\linewidth}
   \includegraphics[width=0.2\linewidth]{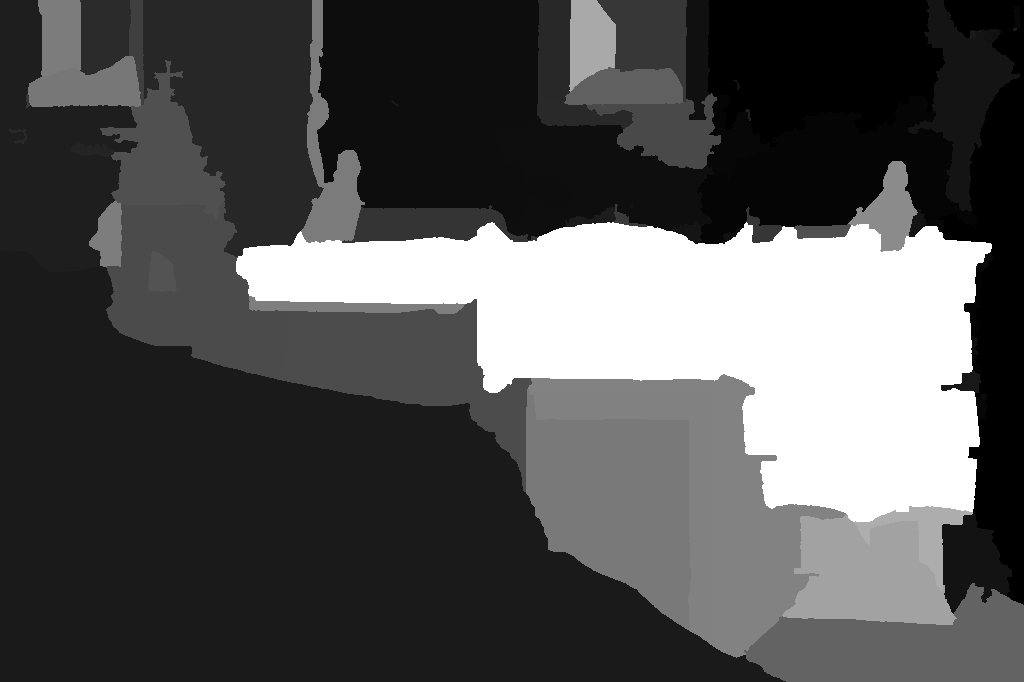}\\\vspace{1mm}
   \includegraphics[width=0.2\linewidth]{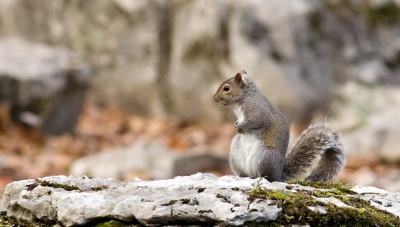}
   \includegraphics[width=0.2\linewidth]{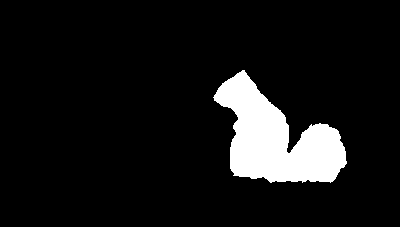}\hspace{0.02\linewidth}
   \includegraphics[width=0.2\linewidth]{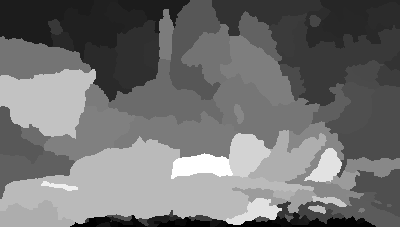}\\\vspace{1mm}
   \includegraphics[width=0.2\linewidth]{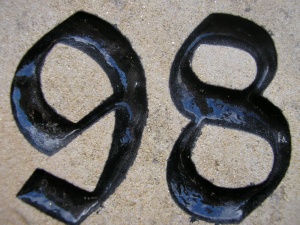}
   \includegraphics[width=0.2\linewidth]{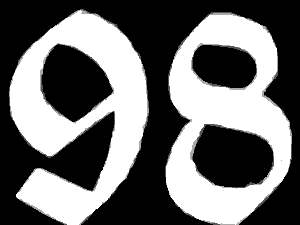}\hspace{0.02\linewidth}
   \includegraphics[width=0.2\linewidth]{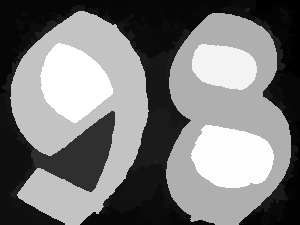}
  \end{center}
  \caption{Typical error cases of PaTS. Left to right: image, ground truth, and resulting saliency map.}
  \label{fig:errors}
 \end{figure}
 Needless to say, our PaTS model also fails at times. Figure~\ref{fig:errors} displays
 some more frequent causes: in the top row example, it merged co-aligned structures to the
 main object. In the second row, the hierarchical clustering was already unreliable. The
 third row shows an example where enclosed background parts became most salient due to only local
 decisions.
 
 Our PaT-Saliency is very efficient to compute and therefore suited to be deployed on a 
 mobile platform. The hierarchical clustering of~\cite{Klein2016} takes about 88 milliseconds
 for a $400\times300$ image (usual MSRA10K size) on a single core of an Intel Xeon X5680 at
 3.33 GHz. Saliency calculation and rendering comes almost for free at only 2 more milliseconds.
 The timings increase only linear in the number of pixels.

\subsection{Overall System Reliability}
\label{subsec:grasp_eval}
 \begin{figure}[t]
  \begin{center}
   \includegraphics[width=0.6\linewidth]{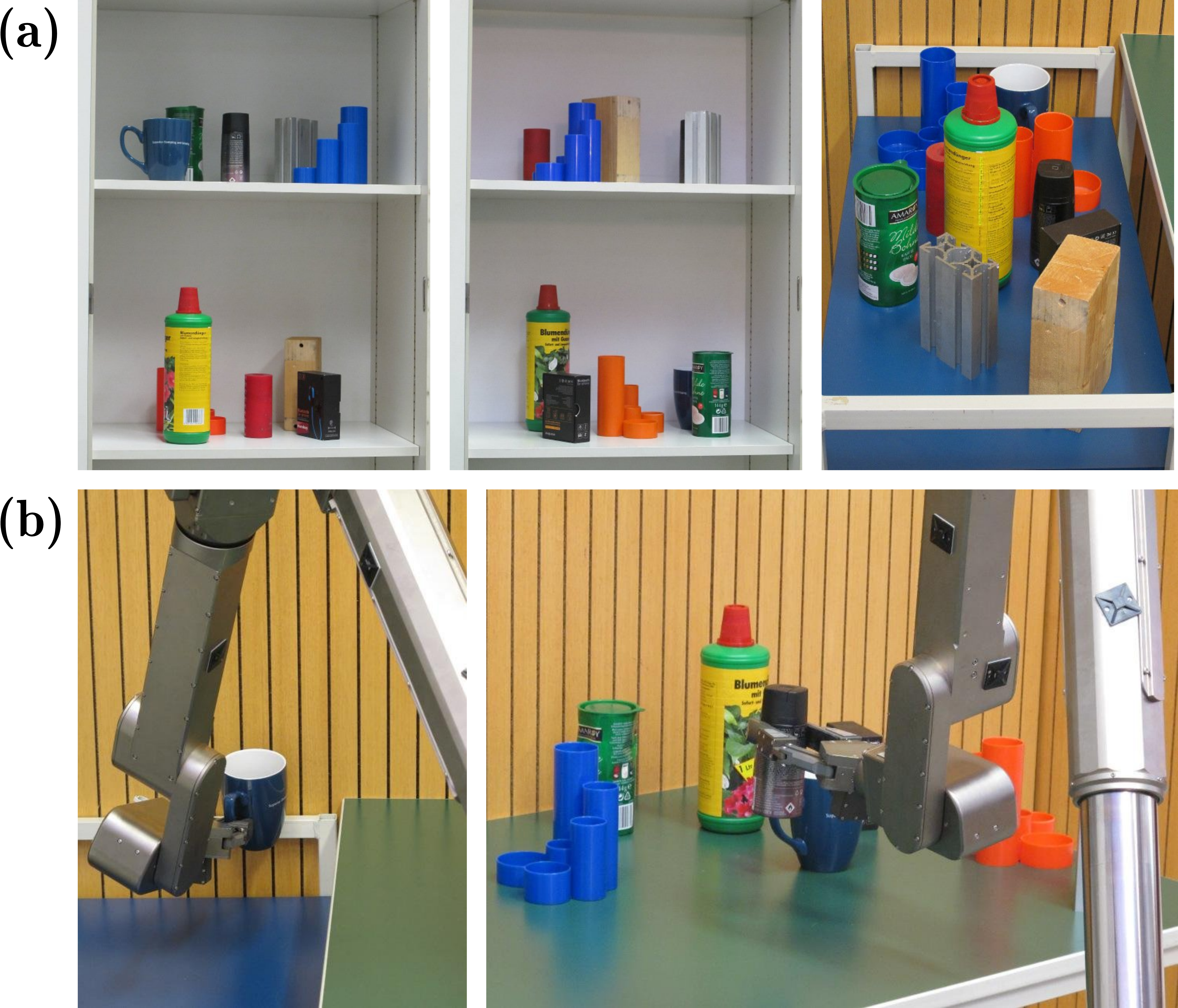}
  \end{center}
  \caption{(a) Start configurations used for the grasping trials with randomly placed set of objects. (b) Example of a pick operation (left) and a place task (right).}
  \label{fig:scenes}
 \end{figure}
To test our saliency based strategy for segmenting unknown objects in a typical pick and place scenario, we also ran a series of trials with a mobile manipulator from an EOD context. The used hardware is a tEODor platform as mobile base with a 7-DOF telemax manipulator mounted on top and a parallel gripper as endeffector. For obtaining the needed depth information we used an ASUS Xtion RGB-D camera on a pan-tilt unit. A laptop served as remote control station for the operator and was used to run the graphical user interface. All other processing was done aboard the robot, equipped with a PC with an Intel i7-2710QE CPU at 2.1 GHz. Segmentation results were available without noticeable delay.

Our grasping experiment contained three sets of trials, trying to pick up and afterwards place 10 objects each: a wooden cuboid, a cardboard packaging box, two pencil holders, a metal profile, a hollow rubber cylinder, a cup, a large bottle, and two differently sized packaging cylinders. The objects were selected to be of varying size, shape, structure, color, and texture. The first two trial sets started with all objects being placed inside a cabinet and the goal was to move all of them to a nearby table. The third trial set was showcasing a tabletop scene, where all objects had to be moved from a lower shelf to the same table as before. The three start configurations used for the trials are shown in Figure~\ref{fig:scenes}a. An exemplary pick and place operation is depicted in Figure~\ref{fig:scenes}b.

To evaluate the effectiveness of our approach we used overall grasp success as measure. From the three sets of trials we obtained data on 30 different grasp attempts, 27 of which successfully worked on first try. In some of those cases though, the operator needed to use the interface to choose a different place position on the table, after having selected a not suitable point at first. This behavior was mostly seen when trying to efficiently use the available space on the table, which at times would have resulted in a collision. In this case the grasp subroutine rejects the supplied input before even attempting to move the manipulator and a new place point has to be selected. Of the remaining three grasp attempts, one had to be repeated and worked on second try (rubber cylinder). One needed a short manual intervention by the operator, moving the turret joint by a few degrees, because the grasped object was too close to the back of the cabinet and the planer got stuck in a collision state (wooden cuboid). The third one resulted in a failed grasp attempt on the cardboard packaging box. This was due to the gripper having passed the box early on while grasping a different object and due to inaccuracies slightly colliding with the box and knocking it over. The box ended up lying flat on the cabinet shelf with the largest side facing the front and was not graspable with the possible gripper aperture.

\begin{table}[t]
\caption{Results of our grasping trials.}
\label{table:grasp-experiments}
\begin{center}
\begin{tabular}{| l || c | c | c |}
\hline
object & cabinet 1 & cabinet 2 & tabletop \\
\hline \hline
wooden cuboid & (\cmark) & \cmark & \cmark \\
\hline
cardboard box & \xmark & \cmark & \cmark \\
\hline
blue pen holder & \cmark & \cmark & \cmark \\
\hline
orange pen holder & \cmark & \cmark & \cmark \\
\hline
metal profile & \cmark & \cmark & \cmark \\
\hline
rubber cylinder & \cmark & \cmark & (\cmark) \\
\hline
cup & \cmark & \cmark & \cmark \\
\hline
bottle & \cmark & \cmark & \cmark \\
\hline
green cylinder & \cmark & \cmark & \cmark \\
\hline
dark cylinder & \cmark & \cmark & \cmark \\
\hline
\end{tabular}
\end{center}
\end{table}

Table  \ref{table:grasp-experiments}
shows a detailed summary of the conducted trials and their outcome. Please also refer to the accompanying video to see the segmentation approach and the corresponding grasp attempt for several objects.

\section{Conclusions and Future Work}

We proposed PaTS, a very efficient transform of an image hierarchical clustering into
a likewise hierarchical saliency function, which was shown to be particularly suited
for salient object detection. Besides boasting state-of-the-art performance on web
images, it proved applicable for camera images from a mobile robot as well.
Obtaining a very accurately segmented object aids in finding a good grasp point
and the proper collision box.
Therefore, using our segmentation approach in conjunction with a simple grasping method
empowers the operator to pick and place objects even in non-tabletop scenes such as unpacking a cabinet.

In the future, we want to integrate an automatic grasp point estimation to our system
in order to further simplify the operation. Moreover, we want to transfer saliency from
single images to 3D maps, possibly built from multiple views.


\bibliographystyle{abbrv}
\bibliography{references}

\end{document}